\documentclass{article}

\usepackage{arxiv}

\usepackage{amsmath, amsfonts, bm, amsthm}
\usepackage{algorithm}
\usepackage{caption}
\usepackage{algpseudocode}
\usepackage{subcaption}
\usepackage{graphicx}
\usepackage{hyperref}
\usepackage{url}
\usepackage[table]{xcolor}

\def\eqref#1{equation~\ref{#1}}

\def\1{\bm{1}}

\def\rvw{{\mathbf{w}}}
\def\rvx{{\mathbf{x}}}

\def\rvz{{\mathbf{z}}}

\def\vs{{\bm{s}}}

\def\vx{{\bm{x}}}

\def\mI{{\bm{I}}}

\DeclareMathAlphabet{\mathsfit}{\encodingdefault}{\sfdefault}{m}{sl}
\SetMathAlphabet{\mathsfit}{bold}{\encodingdefault}{\sfdefault}{bx}{n}

\newcommand{\pdata}{p_{\rm{data}}}

\newcommand{\ptargetNP}{p_{\rm{target}}^{\theta}}
\newcommand{\qtargetNP}{q_{\rm{target}}}

\newcommand{\E}{\mathbb{E}}

\newcommand{\R}{\mathbb{R}}

\newtheorem{theorem}{Theorem}
\newtheorem{lemma}{Lemma}
\newtheorem{proposition}{Proposition}

\DeclareMathOperator*{\argmax}{arg\,max}
\DeclareMathOperator*{\argmin}{arg\,min}

\usepackage[utf8]{inputenc} 
\usepackage[T1]{fontenc}    
\usepackage{hyperref}       
\usepackage{url}            
\usepackage{booktabs}       
\usepackage{amsfonts}       
\usepackage{nicefrac}       
\usepackage{microtype}      
\usepackage{lipsum}		
\usepackage{graphicx}
\usepackage{natbib}
\usepackage{doi}
\usepackage[normalem]{ulem}
\usepackage{cancel, xcolor}
\usepackage{multirow}
\usepackage{pbox}

\title{From Function to Distribution Modeling: \\A PAC-Generative Approach to Offline Optimization}

\author{
    Qiang Zhang\thanks{Q.~Zhang and R. Zhou should both be considered first authors.}\\
    Department of Electrical and Computer Engineering\\
    Texas A\&M University\\
    College Station, TX 77843\\
    \And
    Ruida Zhou\footnotemark[1]\\
    Department of Electrical and Computer Engineering\\
    University of California, Los Angeles\\
    Los Angeles, CA 90095\\
	\AND
	Yang Shen \\
    Department of Electrical and Computer Engineering\\
    Texas A\&M University\\
    College Station, TX 77843\\
	\And
	Tie Liu \\
    Department of Electrical and Computer Engineering\\
    Texas A\&M University\\
    College Station, TX 77843\\
}

\date{}

\renewcommand{\shorttitle}{From Function to Distribution Modeling: A PAC-Generative Approach to Offline Optimization}

\hypersetup{
pdftitle={From Function to Distribution Modeling: A PAC-Generative Approach to Offline Optimization},
pdfsubject={cs.LG, math.OC},
pdfauthor={Qiang Zhang},
pdfkeywords={Offline optimization, Distribution modeling, Generative model, End-to-end learning},
}

\begin{document}
\maketitle

\begin{abstract}
	This paper considers the problem of offline optimization, where the objective function is unknown except for a collection of ``offline" data examples. While recent years have seen a flurry of work on applying various machine learning techniques to the offline optimization problem, the majority of these work focused on learning a surrogate of the unknown objective function and then applying existing optimization algorithms. While the idea of modeling the unknown objective function is intuitive and appealing, from the learning point of view it also makes it very difficult to tune the objective of the learner according to the objective of optimization. Instead of learning and then optimizing the unknown objective function, in this paper we take on a less intuitive but more direct view that optimization can be thought of as a process of sampling from a generative model. To learn an effective generative model from the offline data examples, we consider the standard technique of ``re-weighting", and our main technical contribution is a probably approximately correct (PAC) lower bound on the natural optimization objective, which allows us to jointly learn a weight function and a score-based generative model. The robustly competitive performance of the proposed approach is demonstrated via empirical studies using the standard offline optimization benchmarks.
\end{abstract}


\section{Introduction}\label{sec:intro}
Offline optimization refers to the problem of optimizing an \emph{unknown} real-valued objective function $f$ based only on a collection of ``offline" data examples $\left ( (\vx_i,f(\vx_i)):i \in [m] := \{1,2,\ldots,m\} \right )$, where each $\vx_i$ is an independent sample drawn from an \emph{unknown} data-generating distribution $\pdata$. Aside from these examples, no additional information on the objective function $f$ is available prior to or during the optimization process, and hence the name ``offline optimization". 
This rather restrictive setting is particularly relevant to the optimization scenarios where: i) the objective function is very complex and no structural information is available; and ii) querying the objective function is very expensive. 
Potential applications include the design of proteins \citep{kolli2022datadriven}, chemical molecules \citep{gomez2018automatic}, DNA sequences \citep{killoran2017generating}, aircrafts \citep{hoburg2014geometric}, robots \citep{liao2019data}, and hardware accelerators \citep{kumar2022datadriven}. 

Obviously, offline optimization is a much more challenging setting than standard optimization \citep{beck2017first} (where \emph{full} structural information on the objective function is available) or black-box optimization \citep{Audet2017} (where even though no \emph{structural} information on the objective function is available, the objective function can be queried upon during the optimization process). Therefore, instead of aiming at the global optima, for offline optimization we are usually satisfied with finding a few candidates, among which there are \emph{significantly} better solutions than the existing offline observations.

Traditionally, offline optimization has been mainly approached through the \emph{Bayesian} view, i.e., by endowing the unknown objective function $f$ a \emph{prior} distribution. This has led to a large body of work under the name \emph{Bayesian optimization}; see \citet{fu2021offline} and the references therein for the recent progress in this direction. Motivated by the rapid progress in machine learning, recent years have also seen a flurry of work on offline optimization from a \emph{frequentist's} view \citep{brookes2019conditioning, gupta2019feedback,kumar2020model, trabucco2021conservative}, i.e., by modeling the objective function $f$ as a \emph{deterministic but unknown} function. However, most of these work have been focusing on learning a surrogate of the unknown objective function and then applying existing optimization algorithms. Prime examples include \citet{trabucco2021conservative}, which focused on adapting the \emph{gradient} method to the offline setting, and \citet{brookes2019conditioning, gupta2019feedback}, which focused on adapting the \emph{cross-entropy} method \citep{rubinstein1999cross} to the offline setting. While the idea of modeling the unknown objective function is intuitive and appealing, from the learning point of view it also makes it very difficult to tune the objective of the learner according to the objective of optimization \citep{trabucco2021conservative,brookes2019conditioning, gupta2019feedback}. As a result, it is very difficult to gauge whether these previous approaches actually come with any theoretical guarantees. 

In this paper we take on a less intuitive but more \emph{direct} view of optimization and consider it as a process of \emph{sampling} from a \emph{generative model}. There are two natural advantages to this view. First, through sampling \emph{exploration} is now intrinsic in the optimization process. Second, this view allows us to shift our focus from modeling the objective function to modeling a \emph{target distribution}. Unlike learning a surrogate on the objective function, as we shall show, the objective of learning a target distribution can be \emph{naturally} aligned with the objective of optimization, thus bringing \emph{theoretical guarantees} on the optimization performance.

Let $\ptargetNP$ be a \emph{generative model} from which sampling can produce, with high probability, samples whose objective values are significantly better than the offline observations. Note that unlike the traditional generative models, which are trained to generate samples that are ``similar" to the training examples, the goal of our generative model $\ptargetNP$ is to generate samples with \emph{superior} objective values than the offline observations. Relative to the data-generating distribution $\pdata$, these targeted samples with superior objective values are the ``outliers". Therefore, from the learning perspective, our main challenge here is to learn a generative model that generates \emph{outliers} rather than the norm.

To facilitate the learning of a desired generative model, in this paper we shall consider the standard technique of ``re-weighting". Roughly speaking, we shall consider a \emph{weight} function that assigns higher weights to the domain points with higher objective values and then train a generative model using the \emph{weighted} offline examples. This  helps to tune the generative model towards generating samples with high objective values.

Formally, let
\begin{align}\label{eq:q_target}
    \qtargetNP(\vx) = \tilde{w}(f(\vx))\pdata(\vx)
\end{align}
be a \emph{hypothetical} target distribution,
where $\tilde{w}$ is a normalized, non-negative \emph{weight} function such that $\mathbb{E}_{\rvx\sim \pdata}[\tilde{w}(f(\rvx))]=1$, and $\pdata$ is the (unknown) data-generating distribution from which the offline observations $\vx_{[m]}:=(\vx_i:i\in[m])$ were drawn. In our approach, the hypothetical target distribution $\qtargetNP$ plays \emph{dual} roles: On one hand, it serves as the \emph{hypothetical} learning target of the generative model $\ptargetNP$; on the other hand, it is also connected to the unknown data-generating distribution $\pdata$ via the normalized weight function $\tilde{w}$ and hence allows a generative model $\ptargetNP$ to be learned from the offline data examples. Operationally, we would like to train a generative model $\ptargetNP$ such that
$\ptargetNP \approx \qtargetNP$. But what would be a suitable choice for the normalized weight function $\tilde{w}$?

Intuitively, there are two considerations for selecting a normalized weight function $\tilde{w}$. On one hand, from the \emph{utility} point of view, we would like to choose $\tilde{w}$ such that the hypothetical target distribution $\qtargetNP$ focuses most of its densities on the domain points with \emph{superior} objective values. This can be achieved, for example, by choosing $\tilde{w}$ to be heavily \emph{skewed} towards superior objective values. On the other hand, from the \emph{learning} viewpoint, the generative model $\ptargetNP$ is learned from the offline observations, which were generated from the unknown data-generating distribution $\pdata$. If $\tilde{w}$ is chosen to be heavily skewed, the hypothetical target distribution $\qtargetNP$ then becomes very \emph{different} from the data-generating distribution $\pdata$. In this case, learning the generative model $\ptargetNP$ from the offline data examples may be subject to very high \emph{sample} complexity. 

Given these two seemingly \emph{conflicting} considerations, it is natural to make the normalized weight function $\tilde{w}$ (and hence the hypothetical target distribution $\qtargetNP$) to be \emph{learnable} as well. Assume without loss of generality that our goal is to \emph{maximize} the (unknown) objective function $f$. From the optimization point of view, a \emph{natural} optimization objective for identifying a desired generative model $\ptargetNP$ is by maximizing the expected objective value:
\begin{equation}\label{eq:obj}
    J_{\text{opt}}(\theta)=\E_{\rvx\sim \ptargetNP}[f(\rvx)].
\end{equation}
The above objective, however, \emph{cannot} be evaluated for any given model parameter $\theta$, because the objective function $f$ is \emph{unknown}. Instead of trying to learn a surrogate on $f$ (and then use it to guide the training of the generative model), in this paper we shall follow the more traditional learning-theoretic approach of constructing a \emph{probably approximately correct (PAC)} lower bound on the natural optimization objective $J_{\text{opt}(\theta)}$. Unlike $J_{\text{opt}(\theta)}$, the PAC lower bound depends on both $\theta$ and the normalized weight function $\tilde{w}$. As we shall see, not only it captures \emph{both} the aforementioned utility and learnability considerations for selecting $\tilde{w}$, it will also \emph{naturally} suggest a objective, from $\theta$ and $\tilde{w}$ can be \emph{jointly} learned. 


The rest of the paper is organized as follows. In Section \ref{sec:prelim}, we introduce a few technical results for establishing the PAC lower bound on $J_{\rm{opt}}$. The PAC lower bound is  formally presented in Section \ref{sec:main result}. In Section \ref{sec:algo}, we discuss how to leverage the PAC lower bound for jointly learning a weight function and a \emph{score-based} generative model. In Section~\ref{sec:experiment}, we demonstrate the legitimacy and the robustly competitive performance of the proposed learner, first through a toy example and then through the standard offline optimization benchmark datasets. Finally in Section~\ref{sec:related work}, we conclude the paper with an in-depth discussion on the contribution of this paper in the context of several related work. 

\section{Preliminaries}\label{sec:prelim}
In this section, we introduce a few technical results that are essential for constructing the desired PAC lower bound. 

\subsection{Wasserstein distance}
Let $\mu$ and $\nu$ be two probability distributions on $\R^d$. A \emph{coupling} $\gamma$ between $\mu$ and $\nu$ is a joint distribution on $\R^d \times \R^d$ whose marginals are $\mu$ and $\nu$. The $p$-Wasserstein distance between $\mu$ and $\nu$ (with respect to the Euclidean norm) is given by:
\begin{equation}
    W_p(\mu,\nu) =  \left( \inf_{\gamma \in \Gamma(\mu,\nu)} \E_{(\rvx,\Tilde{\rvx}) \sim \gamma} \left[\| \rvx-\Tilde{\rvx} \|^{p} \right]\right)^{1/p},
\end{equation}
where $\Gamma(\mu,\nu)$ is the set of all couplings between $\mu$ and $\nu$, and $\|\cdot\|$ denotes the standard Euclidean norm. 

The $1$-Wasserstein distance, also known as the \emph{earth mover's distance}, has an important equivalent representation that follows from the \emph{duality} theorem of Kantorovich-Rubenstein \citep{Ambrosio2021}: 
\begin{equation}
    W_1(\mu,\nu) = \frac{1}{K} \sup_{\|\tilde{f}\|_{\text{Lip}} \leq K} \left\{\mathbb{E}_{\rvx \sim \mu} [\tilde{f}(\rvx)] - \mathbb{E}_{\rvx \sim \nu} [\tilde{f}(\rvx)]\right\},
\end{equation}
where $\|\cdot\|_{\text{Lip}}$ denotes the \emph{Lipschitz} norm. In our construction, this \emph{dual} representation of the $1$-Wasserstein distance serves as the bridge between the \emph{objective-specific} generative loss and the \emph{generic} generative loss. By the standard Jensen's inequality, we also have
\begin{equation}\label{eq:W1_W2}
    W_1(\mu,\nu) \leq W_2(\mu,\nu)
\end{equation}
for any two distributions $\mu$ and $\nu$. As we shall see, this simple relationship between the $1$-Wasserstein and $2$-Wasserstein distances can help to further connect the \emph{objective-specific} generative loss to the \emph{denoising score matching loss} for training a \emph{score-based} generative model.

\subsection{Denoising diffusion probabilistic model}
\label{sec:score}
In this paper, we shall mainly focus on training a  \emph{score-based} model $\ptargetNP$ such that $\ptargetNP \approx \qtargetNP$. This is motivated by the following connection between the \emph{score function} of the hypothetical target distribution $\qtargetNP$ and the \emph{gradient} of the unknown objective function $f$:
\begin{align}
    \vs_{\text{target}}(\vx) &= \nabla \log \qtargetNP (\vx)= \nabla \log\left[\tilde{w}(f(\vx)) \pdata(\vx)\right] \nonumber \\
    &= \vs_{\text{data}}(\vx) + \frac{\tilde{w}^{'}(f(\vx))}{\tilde{w}(f(\vx))} \nabla f(\vx),\label{eq:score}
\end{align}
where $\vs_{\text{target}}$ and $\vs_{\text{data}}$ are the score functions of $\ptargetNP$ and $\pdata$, respectively. 
If $\tilde{w}$ is \emph{monotone increasing}, the derivative $\tilde{w}^{'}(f(\vx))>0$ for all $\vx\in\mathcal{X}$. In this case, sampling along the direction of $\vs_{\text{target}}$ can \emph{naturally} produce samples with higher objective values.

We are particularly interested in the \emph{denoising diffusion probabilistic model (DDPM)} \citep{song2020score,ho2020denoising} due to its stability and performance on high-dimensional datasets. Here we recall a few essential results on the DDPM. Consider a \emph{forward} process of continuously injecting white Gaussian noise into a signal $\rvx_t$:
\begin{equation}\label{eq:fp}
    d\rvx_t = -\frac{1}{2} \beta(t) \rvx_t dt + \sqrt{\beta(t)} d\rvw_t, \ \ t \in [0, 1],
\end{equation}
where $\beta: [0,1]\rightarrow \mathbb{R}_{++}$ is a positive \emph{noise scheduler}, $\rvw_t$ is a standard \emph{Wiener} process, and time in this process is assumed to flow in the \emph{forward} direction from $t=0$ to $t=1$. Denote by $q_t$ the marginal distribution of $\rvx_t$ from the forward process (\ref{eq:fp}). DDPM is mainly motivated by the fact that the marginal distributions $q_t$, $t\in[0,1]$, can be \emph{recovered} through the following \emph{reverse} process \citep{anderson1982reverse}:
\begin{equation}\label{eq:bp}
    d\rvx_t = - \beta(t)\left(\frac{1}{2}\rvx_t +\vs^\theta_t(\rvx_t) \right) dt + \sqrt{\beta(t)} d\Bar{\rvw}_t,
\end{equation}
where $\vs^\theta_t$ is a model of the score function of $q_t$ and $\Bar{\rvw}_t$ is (again) a standard \emph{Wiener} process but with time flowing \emph{backward} from $t=1$ to $t=0$. More specifically, let $p^\theta_t$ be the marginal distribution of $\rvx_t$, $t\in[0,1]$ from the reverse process (\ref{eq:bp}). If we let $p_1^\theta=q_1$ and $\vs^\theta_t$ be the \emph{exact} score function of $q_t$ for all $t\in[0,1]$, we have $p^\theta_t=q_t$ for \emph{all} $t\in[0,1)$ \citep{bogachev2022fokker}. In particular, the initial distribution of the forward process $q_0$ can be recovered at the end of the reverse process via the score functions of $q_t$, $t\in[0,1]$. Thus, to learn the initial distribution of the forward process $q_0$, it suffices to learn a model $\vs^{\theta}_t$ that approximates the score functions of $q_t$ for all $t\in[0,1]$.

There are several methods \citep{hyvarinen2005estimation,vincent2011connection,song2020sliced} that allow a model $\vs^{\theta}_t$ to be learned from a training dataset drawn from $q_0$. Here we focus on the \emph{denoising score matching method} due to its scalability to large datasets. For the denoising score matching method, a model $\vs^\theta_t$ is learned by minimizing the following \emph{denoising score matching loss}: 
\begin{equation}
    \mathcal{L}_{\text{DSM}}(\theta;q_0) 
    = \E_{\rvx \sim q_0} \left[\ell_{\text{DSM}}^{\theta}(\rvx)\right],
\end{equation}
where 
\begin{align} \label{eq:point_dsm_loss}
    \ell_{\text{DSM}}^{\theta}(\rvx) = \int_{0}^{1} \lambda(t)  \E_{\rvz \sim \mathcal{N}(0, \mI)}  \left[\left \| \vs_t^\theta \left(\rvx_t\right) + \frac{\rvz}{\sigma(t)}  \right \|^2  \right] dt
\end{align} 
is the \emph{point-wise denoising score matching loss}, $\lambda: [0,1]\rightarrow \mathbb{R}_{++}$ can be any positive function, 
\begin{align}\nonumber
    \rvx_t = \sqrt{1-\sigma(t)^2}\rvx+ \sigma(t)\rvz,
\end{align}
and 
\begin{align}\nonumber
    \sigma(t) = \sqrt{1-\exp\left[-\int_0^t \beta(s) ds\right]}.
\end{align}
While the previous denoising score matching loss can be easily estimated from a dataset drawn from $q_0$ (and hence is very \emph{conductive} to learning), \emph{a priori} it is unclear how it would connect to any \emph{generative} loss between $q_0$ and $p_0^\theta$. Interestingly, it was shown in Theorem~2 and Corollary~3 of \citet{kwon2022scorebased} that under some (relatively) mild conditions\footnote{The readers are referred to Section~3.1 of \citet{kwon2022scorebased} for the assumptions under which the inequality (\ref{eq:Secret}) holds.
} on $\beta$, $q_0$, and $\vs_\theta$, by choosing $\lambda(t)=\beta(t)$ we have
\begin{align}
\label{eq:Secret}
    W_2(q_0,p_0^\theta) &\leq c_0\sqrt{\mathcal{L}_{\text{DSM}}(\theta;q_0)}+c_1W_2(q_1,p_1)\nonumber\\
    &=c_0\sqrt{\E_{\rvx \sim q_0} \left[ \ell_{\text{DSM}}^{\theta}(\rvx) \right]}+c_1W_2(q_1,p_1),
\end{align}
where $c_0$ and $c_1$ are constants that only depend on the choice of the noise scheduler $\beta$ and some prior knowledge on $p_0$ and $\vs_t^\theta$ but is independent of the model parameter $\theta$. In \citet{kwon2022scorebased}, this result was coined as ``score-based generative modeling \emph{secretly} minimizes the Wasserstein distance".

For our purposes, let $q_0=\qtargetNP$ and $p_1^\theta$ be the standard Gaussian distribution $\mathcal{N}$. If we denote $q_1$ and $p_0^\theta$ by $\bar{q}_{\text{target}}$ and $\ptargetNP$ respectively, we have
\begin{align}
    W_2(\qtargetNP,\ptargetNP) \leq c_0\sqrt{\E_{\rvx \sim \qtargetNP} \left[\ell_{\text{DSM}}^{\theta}(\rvx) \right]}+c_1W_2(\bar{q}_{\text{target}},\mathcal{N}),\label{eq:Secret2}
\end{align}
where $c_0$ and $c_1$ are constants that only depend on the choice of the noise scheduler $\beta$ and some prior knowledge on $\pdata$, $\tilde{w}$, and $\vs_t^\theta$; otherwise, they are independent of the model parameter $\theta$ and the choice of the normalized weight function $\tilde{w}$. We mention here that the output distribution of the forward process $\bar{q}_{\text{target}}$ is potentially dependent on the choice of $\tilde{w}$, even though this dependency is not explicit from the notation. In practice, $\bar{q}_{\text{target}}$ can be made very close to the standard Gaussian distribution $\mathcal{N}$ with an appropriate choice of the noise scheduler $\beta$. Therefore, the Wasserstein distance $W_2(\bar{q}_{\text{target}},\mathcal{N})$ is very small and is usually disregarded from the learning process.


\subsection{Generalization bound for weighted learning}
Let $\ell_\theta: \mathcal{X}\rightarrow \mathbb{R}$ be a \emph{bounded} loss function parameterized by $\theta\in\Theta$ such that $0 \leq \ell_\theta(\vx) \leq \Delta$ for all $\vx \in \mathcal{X}$ and all $\theta\in\Theta$. Consider the problem of estimating the expected \emph{weighted} loss $\mathcal{L}_p(\theta,\tilde{w})=\mathbb{E}_{\rvx\sim p}[\tilde{w}(\rvx)\ell_\theta(\rvx)]$, where $\tilde{w}:\mathcal{X}\rightarrow \mathbb{R}_+$ is a normalized, \emph{bounded} weight function such that $\mathbb{E}_{\rvx\sim p}[\tilde{w}(\rvx)]=1$ and $0 \leq \tilde{w}(\vx) \leq B$ for all $\vx \in \mathcal{X}$. We have the following PAC upper bound, with respect to the parameter family $\Theta$, on the expected weighted loss $\mathcal{L}_p(\theta,\tilde{w})$ for any given $\tilde{w}$.

\begin{lemma}\label{lemma}
    For any given $\tilde{w}$, with probability $\geq 1-\delta$ we have for \emph{any} $\theta \in \Theta$
    \begin{align}
    \mathcal{L}_p(\theta,\tilde{w}) \leq \hat{\mathcal{L}}_{\vx_{[m]}}(\theta,\tilde{w})+2\Delta\sqrt{\hat{V}_{\vx_{[m]}}(\tilde{w})}+2\hat{\mathfrak{R}}_{\vx_{[m]}}(\Theta) + 3 \sqrt{ \frac{ 2 B \Delta 
    \log(2/\delta) }{m} }, \label{eq:Lemma1}
\end{align}
where 
\begin{align}\nonumber
    \hat{\mathcal{L}}_{\vx_{[m]}}(\theta,\tilde{w})=\frac{1}{m}\sum_{i=1}^m\tilde{w}(\vx_i)\ell_\theta(\vx_i)
\end{align}
is the \emph{empirical} weighted loss over the training dataset $\vx_{[m]}$, 
\begin{align}\nonumber
    \hat{V}_{\vx_{[m]}}(\tilde{w})=\frac{1}{m}\sum_{i=1}^m\left(\tilde{w}(\vx_i)-1\right)^2
\end{align}
is the \emph{empirical} variance of $\tilde{w}$ over $\vx_{[m]}$, and 
\begin{align}\nonumber
    \hat{\mathfrak{R}}_{\vx_{[m]}}(\Theta)=\E_{\boldsymbol{\sigma}_{[m]}}\left[ \sup_{\theta \in \Theta} \frac{1}{m} \sum_{i=1}^m \sigma_i \ell_\theta(\vx_i)\right]    
\end{align}
is the \emph{empirical} Rademacher complexity with respect to the parameter family $\Theta$ over $\vx_{[m]}$.
\end{lemma}

A proof of the above lemma is deferred to Appendix~\ref{sec:lemma proof} to enhance the flow of the paper. The main insight from the above lemma is that the \emph{generalization error} (with respect to the parameter $\theta$) between the expected weighted loss $\mathcal{L}_p(\theta,\tilde{w})$ and the empirical weighted loss $\hat{\mathcal{L}}_{\vx_{[m]}}(\theta,\tilde{w})$  can be controlled by controlling the \emph{complexity} of the model class $\Theta$ and the \emph{variance} of the normalized weight function $\tilde{w}$.

\section{Main Results}\label{sec:main result}
We first introduce a \emph{distribution-dependent} surrogate on the natural optimization objective $J_{\text{opt}}(\theta)$ for a \emph{score-based} generative model $\ptargetNP$.

\begin{proposition} \label{prop:score}
    Assume that the unknown objective function $f$ is $K$-Lipschitz and the generative model $\ptargetNP$ is a DDPM. Under the assumptions from Section~3.1 of \citet{kwon2022scorebased} on the noise scheduler $\beta$, the data-generating distribution $\pdata$, the normalized weight function $\tilde{w}$, and the score-function model $\vs_t^\theta$, we have 
    \begin{align}
        J_{\text{opt}}(\theta) \geq \underbrace{\mathbb{E}_{\rvx \sim \pdata} \left[
        \tilde{w}(f(\rvx))f(\rvx)\right]}_{\text{expected utility}}- \underbrace{c_0K\sqrt{\mathbb{E}_{\rvx \sim \pdata}\left[\tilde{w}(f(\rvx))\ell_{\text{DSM}}^{\theta}(\rvx)\right]}}_{\text{expected generative loss}}-c_1KW_2(\bar{q}_{\text{target}
        },\mathcal{N}), \label{eq:Surro1}
    \end{align}
    where $\ell_{\text{DSM}}^{\theta}$ is the point-wise denoising score matching loss of $\vs_\theta$ as defined in (\ref{eq:point_dsm_loss}), $\bar{q}_{\text{target}}$ is the output distribution of the forward process (\ref{eq:fp}), $\Phi$ is the standard Gaussian distribution, and $c_0$ and $c_1$ are constants that are independent of the model parameter $\theta$ and $\tilde{w}$.
\end{proposition}

\begin{proof}
    We start by writing $J_{\text{opt}}(\theta)$ as:
    \begin{align}
    J_{\text{opt}}(\theta)=\mathbb{E}_{\rvx \sim \qtargetNP}[f(\rvx)]-\left\{\mathbb{E}_{\rvx \sim \qtargetNP} [f(\rvx)]-\mathbb{E}_{\rvx \sim \ptargetNP} [f(\rvx)]\right\}.\label{eq:dist}
    \end{align}
    By the definition of $\qtargetNP$ from (\ref{eq:q_target}), we have
    \begin{align}
        \mathbb{E}_{\rvx \sim \qtargetNP} [f(\rvx)]
        &= \mathbb{E}_{\rvx \sim \pdata} \left[
        \tilde{w}(f(\rvx))f(\rvx)\right].\label{eq:Utility}
    \end{align}
    Furthermore,
    \begin{align}
        \mathbb{E}_{\rvx \sim \qtargetNP} [f(\rvx)]-\mathbb{E}_{\rvx \sim \ptargetNP} [f(\rvx)] &\leq \sup_{\|\tilde{f}\|_{\text{Lip}} \leq K} \left\{\mathbb{E}_{\rvx \sim \qtargetNP} [\tilde{f}(\rvx)] - \mathbb{E}_{\rvx \sim \ptargetNP} [\tilde{f}(\rvx)]\right\}\nonumber\\
        &= K\cdot W_1(\qtargetNP,\ptargetNP), \label{eq:W1}
    \end{align}
    where the first inequality follows directly from the assumption that $f$ is $K$-Lipschitz, and the second equality follows from the dual representation (\ref{eq:W1}) of the $1$-Wasserstein distance. 
    
    Under the assumption that $\ptargetNP$ is a DDPM, we can further bound the $1$-Wasserstein distance $W_1(\qtargetNP,\ptargetNP)$ as:
    \begin{align}
        W_1(\qtargetNP,\ptargetNP) \leq  W_2(\qtargetNP,\ptargetNP) \leq  c_0\sqrt{\mathbb{E}_{\rvx \sim \pdata}\left[\tilde{w}(f(\rvx))\ell_{\text{DSM}}^{\theta}(\rvx)\right]}+c_1W_2(\bar{q}_{\text{target}
            },\mathcal{N}),\label{eq:DSM2}
    \end{align}
where the first inequality follows from (\ref{eq:W1_W2}), and the second inequality follows from (\ref{eq:Secret2}) and the definition of $\qtargetNP$ in (\ref{eq:q_target}). 

Substituting (\ref{eq:Utility}), (\ref{eq:W1}), and (\ref{eq:DSM2}) into (\ref{eq:dist}) completes the proof of (\ref{eq:Surro1}). 
\end{proof}

The distribution-dependent surrogate on the right-hand side of (\ref{eq:Surro1}) can be converted into a \emph{PAC} lower bound using the standard \emph{complexity} theory for machine learning. The result is summarized in the following theorem.

\begin{theorem} \label{theorem}
    Assume that: i) the unknown objective function $f$ is $K$-Lipschitz and satisfies $|f(\vx)| \leq F$ for all $\vx \in\mathcal{X}$; ii) the generative model $\ptargetNP$ is a DDPM; iii) the point-wise denoising score matching loss $\ell^{\theta}_{\text{DSM}}$ satisfies $0 \leq \ell^{\theta}_{\text{DSM}}(\vx)\leq \Delta$ for all $\vx \in\mathcal{X}$ and all $\theta\in\Theta$; and iv) the conditions from Section~3.1 of \citet{kwon2022scorebased} on the noise scheduler $\beta$, the data-generating distribution $\pdata$, the normalized weight function $\tilde{w}$, and the score-function model $\vs_t^\theta$ are satisfied. Let $\tilde{\mathcal{W}}$ be the collection of all normalized weight functions $\tilde{w}$ that are $L$-Lipschitz and satisfy $0 \leq \tilde{w}(y) \leq B$ for any $y\in[-F,F]$. With probability $\geq 1-\delta$, we have for \emph{any} $\tilde{w}\in\tilde{\mathcal{W}}$ and \emph{any} $\theta\in \Theta$,
    \begin{align}
        J_{\text{opt}}(\theta) \geq &\underbrace{\hat{J}_{\vx_{[m]}}(\tilde{w})}_{\text{empirical utility}}-\underbrace{c_0K\sqrt{\hat{\mathcal{L}}_{\vx_{[m]}}(\theta,\tilde{w})}}_{\text{empirical generative loss}}- \underbrace{c_0K\sqrt{2\Delta}\sqrt[4]{\hat{V}_{\vx_{[m]}}(\tilde{w})}}_{\text{empirical variance}}-\nonumber\\
        &c_1KW_2(\bar{q}_{\text{target}
        },\mathcal{N})-c_0K\sqrt{2\hat{\mathfrak{R}}_{\vx_{[m]}}(\Theta)}-O\left(1/\sqrt[8]{m}\right), \label{eq:Surro2}
    \end{align}
    where 
    \begin{align}\nonumber
        \hat{J}_{\vx[m]}(\tilde{w})=\frac{1}{m}\sum_{i=1}^m\tilde{w}(f(\vx_i))f(\vx_i)    
    \end{align}
    is the \emph{empirical} utility of $\tilde{w}$, 
    \begin{align}\nonumber
        \hat{\mathcal{L}}_{\vx_{[m]}}(\theta,\tilde{w})=\frac{1}{m}\sum_{i=1}^m\tilde{w}(f(\vx_i))\ell_{\text{DSM}}^\theta(\vx_i)    
    \end{align}
    is the \emph{empirical} weighted denoising score matching loss of $\vs_t^\theta$, 
    \begin{align}\nonumber
        \hat{V}_{\vx_{[m]}}(\tilde{w})=\frac{1}{m}\sum_{i=1}^m\left(\tilde{w}(f(\vx_i))-1\right)^2    
    \end{align}
    is the \emph{empirical} variance of $\tilde{w}$, 
    \begin{align}\nonumber
        \hat{\mathfrak{R}}_{\vx_{[m]}}(\Theta)=\E_{\boldsymbol{\sigma}_{[m]}}\left[ \sup_{\theta \in \Theta} \frac{1}{m} \sum_{i=1}^m \sigma_i \ell_{\text{DSM}}^\theta(\vx_i)\right]    
    \end{align}
    is the \emph{empirical} Rademacher complexity with respect to the parameter family $\Theta$, and the last term $O\left(1/\sqrt[8]{m}\right)$ is independent of the model parameter $\theta$ and $\tilde{w}$.
\end{theorem}

For completeness, here we outline the main steps of the proof. To prove (\ref{eq:Surro2}), let us first fix a $\tilde{w}\in\tilde{\mathcal{W}}$. Given $\tilde{w}$, we can: i) apply the standard Hoeffding's inequality \citep{hoeffding1994probability} to obtain a data-dependent lower bound on  $\mathbb{E}_{\rvx \sim \pdata} \left[\tilde{w}(f(\rvx))f(\rvx)\right]$; and ii) apply Lemma~\ref{lemma} to obtain a PAC upper bound on $\mathbb{E}_{\rvx \sim \pdata}\left[\tilde{w}(f(\rvx))\ell_{\text{DSM}}^{\theta}(\rvx)\right]$ with respect to the parameter family $\Theta$. In light of Proposition~\ref{prop:score}, combining these two bounds gives us a \emph{conditional} PAC lower bound on $J_{\text{opt}}(\theta)$ with respect to the parameter family $\Theta$ for any \emph{fixed} $\tilde{w}$. Finally, we get rid of the conditioning on $\tilde{w}$ via the standard \emph{covering} argument. The details of the proof can be found in Appendix~\ref{sec:theorem proof}.

According to the PAC lower bound (\ref{eq:Surro2}), in order to maximize $J_{\text{opt}}(\theta)$, we need to simultaneously maximize the \emph{utility} of $\tilde{w}$ and minimize the \emph{weighted generative loss} of $\vs_\theta$ and the \emph{variance} of $\tilde{w}$. Therefore, the PAC lower bound (\ref{eq:Surro2}) captures both the utility and learnability considerations for selecting a normalized weight function $\tilde{w}$.

\section{Algorithm}\label{sec:algo}
To jointly learn a normalized function $\tilde{w}$ and a score-function model $\vs_t^\theta$, first note that the last two terms of the PAC lower bound (\ref{eq:Surro2})are independent of $\theta$ and $\tilde{w}$ and hence can be ignored from the learning objective. The forth term is due to the ``initial" sampling error of the reverse process. As discussed previously in Section~\ref{sec:score}, while this term is potentially dependent on the normalized weight function $\tilde{w}$, in practice it can be made very small by choosing an appropriate noise scheduler $\beta$ and hence will be ignored from our learning objective. To make the first three terms 
\emph{learnable}, we consider the following two modifications to the bound. 

First, the coefficients $c_0 K$ and $c_0 K\sqrt{2\Delta}$ in the second and the third term require some prior knowledge on the unknown data-generating distribution $\pdata$ and the unknown objective function $f$. In practice, we replace them by two \emph{hyper-parameters} $\lambda$ and $\alpha$, respectively. We mention here that the hyper-parameter $\alpha$ plays a particular important role in the learning objective, as it controls the utility-learnability tradeoff for selecting a normalized weight function $\tilde{w}$. 

Second, the weight function $\tilde{w}$ needs to be \emph{normalized} with respect to the unknown data-generating distribution $\pdata$ and the unknown objective function $f$. In practice, we let
$\tilde{w}(\cdot) = \frac{w_\phi(\cdot)}{Z_\phi}$, where $w_\phi$ is an \emph{un-normalized} weight function parameterized by a second parameter $\phi$, and $Z_\phi=\mathbb{E}_{\rvx \sim \pdata}[w_\phi(f(\rvx))]$ is the normalizing constant. While the exact calculation of $Z_\phi$ again requires the knowledge of $\pdata$ and $f$, in practice it can be easily estimated from the offline data examples as $\hat{Z}_{\phi} = \frac{1}{m}\sum_{i=1}^mw_{\phi}(f(\vx_i))$.

Incorporating the above changes to the PAC lower bound (\ref{eq:Surro2}) leads to the following objective for jointly learning a (un-normalized) weight function $w_\phi$ and a score-function model $\vs_t^\theta$:
\begin{align}
    J_{\alpha,\lambda}(\theta,\phi) = \frac{1}{m}\sum_{i=1}^m \frac{w_{\phi}(f(\vx_i))f(\vx_i)}{\hat{Z}_{\phi}}- \lambda\sqrt{\frac{1}{m}\sum_{i=1}^m\frac{w_{\phi}(f(\vx_i))\ell_{\text{DSM}}^{\theta}(\vx_i)} {\hat{Z}_{\phi}}}- \alpha \sqrt[4]{\frac{1}{m}\sum_{i=1}^m\left(\frac{w_\phi(f(\vx_i))}{\hat{Z}_{\phi}}-1\right)^2}.\label{eq:Surro3}
\end{align}

To maximize $J_{\alpha,\lambda}$, we consider the standard \emph{alternating} procedure between maximizing over the parameter $\theta$ and maximizing over the parameter $\phi$. One advantage of this strategy is that for a fixed $\phi$, maximizing $J_{\alpha,\lambda}$ over $\theta$ is equivalent to minimizing the weighted denoising score matching loss $\frac{1}{m}\sum_{i=1}^mw_\phi(f(\vx_i))\ell_{\text{DSM}}^{\theta}(\vx_i)$ over $\theta$, which is \emph{additive} with respect to the training examples. Therefore, for a fixed $\phi$, updating $\theta$ can be based on a very efficient \emph{stochastic} estimate of the gradient. On the other hand, for a fixed $\theta$, updating $\phi$ may have to rely on the much more cumbersome task of calculating the \emph{exact} gradient. Fortunately, $w_\phi$ is only a \emph{scalar} function, which makes the gradient calculation with respect to $\phi$ somewhat manageable. The detailed training and optimization procedures can be found in Appendix~\ref{sec:training details}.

\section{Experimental Results}\label{sec:experiment}

\subsection{A toy example}\label{sec:toy example}
\begin{figure}[!t]
    \centering
    \begin{subfigure}[b]{0.4\linewidth}
        \centering
        \includegraphics[width=\linewidth]{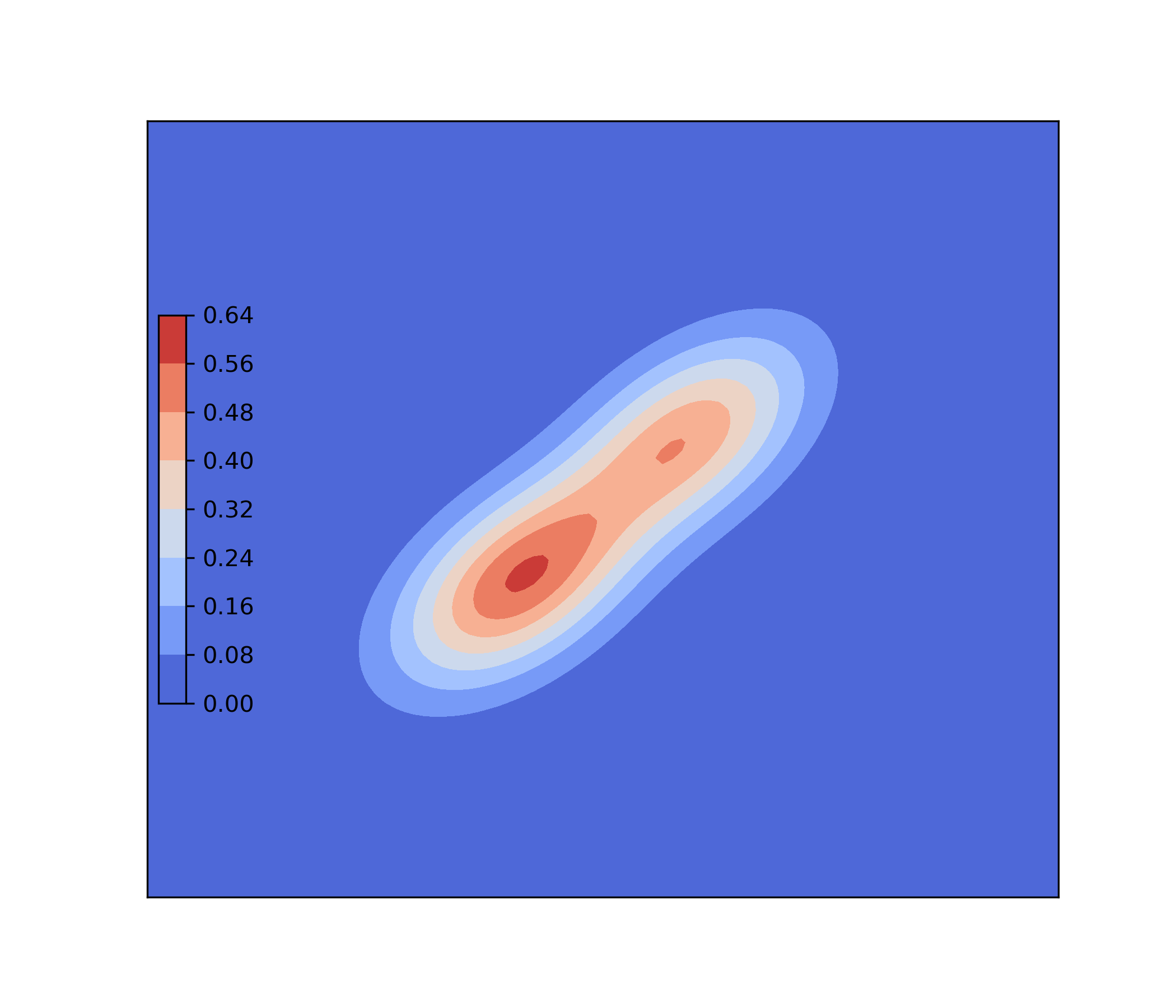}
        \caption{Objective function}
        \label{fig:contour}
    \end{subfigure}
    \hspace{10pt}
    \begin{subfigure}[b]{0.4\linewidth}
        \centering
        \includegraphics[width=\linewidth]{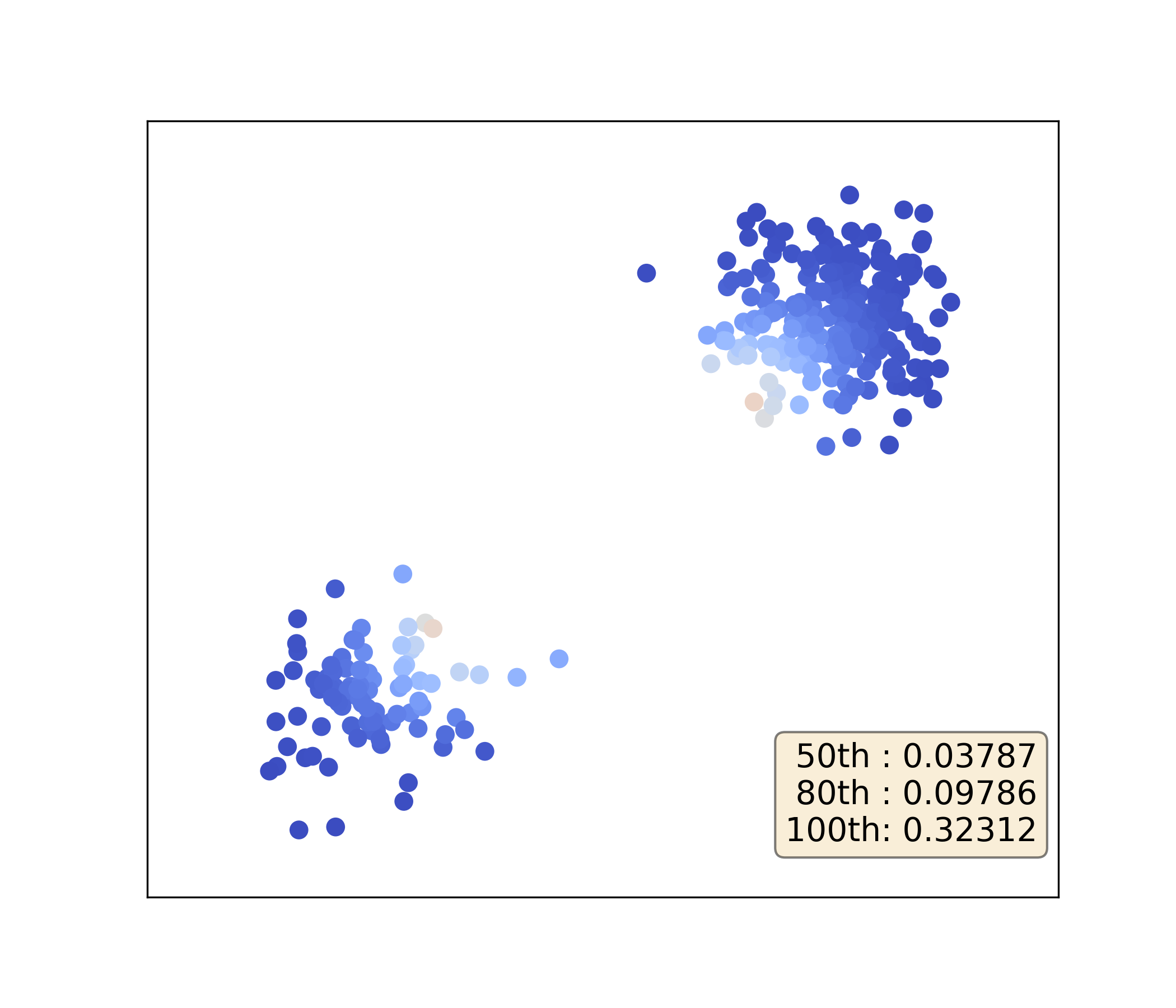}
        \caption{Initial samples}
        \label{fig:initial}
    \end{subfigure}
\caption{A toy example: The objective function and the initial samples.}
\label{fig:toy1}
\end{figure}

We first experimentally validate the proposed learning algorithm using a toy example in $\mathbb{R}^2$. In this example, the unknown objective function $f$ is a mixture of two Gaussian density functions: 
\begin{align}\nonumber
    f(\vx)= 2\sqrt{3}\pi \left[0.45\cdot\mathcal{N}(\vx;\boldsymbol{\mu}_1,\boldsymbol{\Sigma})+ 0.55\cdot\mathcal{N}(\vx;\boldsymbol{\mu}_2,\boldsymbol{\Sigma})\right],
\end{align} 
where $\boldsymbol{\mu}_1=[1.5,1.5]^t$, $\boldsymbol{\mu}_2=[-1.5,-1.5]^t$, $\boldsymbol{\Sigma}=\left[
\begin{array}{cc}
    2 & 1 \\
    1 & 2
\end{array}\right]$, and the unknown data-generating distribution $\pdata$ is a mixture of two Gaussian distributions: 
\begin{align}\nonumber
    \pdata(\vx)=0.3\cdot\mathcal{N}(\vx;\boldsymbol{\mu}_3,\mI)+ 0.7\cdot\mathcal{N}(\vx;\boldsymbol{\mu}_4,\mI),    
\end{align}
where $\boldsymbol{\mu}_3=[-4,-4]^t$, $\boldsymbol{\mu}_4=[4,4]^t$, and $\mI$ is the $2\times2$ identity matrix. The weight function $w_\phi$ and the score-function model $\vs_\theta$ are jointly learned by maximizing the proposed objective (\ref{eq:Surro3}). 

\begin{figure}[!ht]
    \centering
    \begin{subfigure}[b]{0.32\linewidth}
        \centering
        \includegraphics[width=\linewidth]{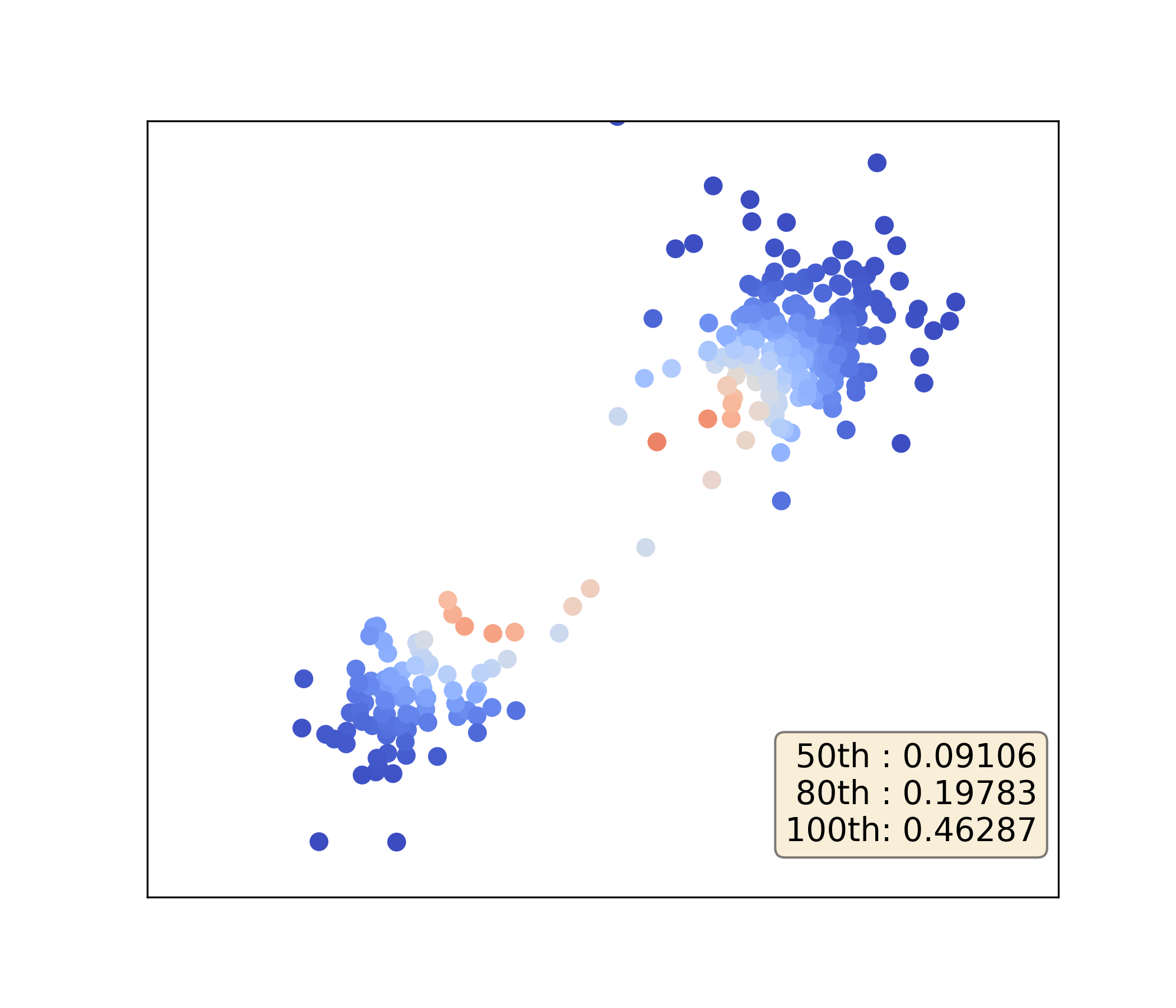}
    \end{subfigure}
    \hfill
    \begin{subfigure}[b]{0.32\linewidth}
        \centering
        \includegraphics[width=\linewidth]{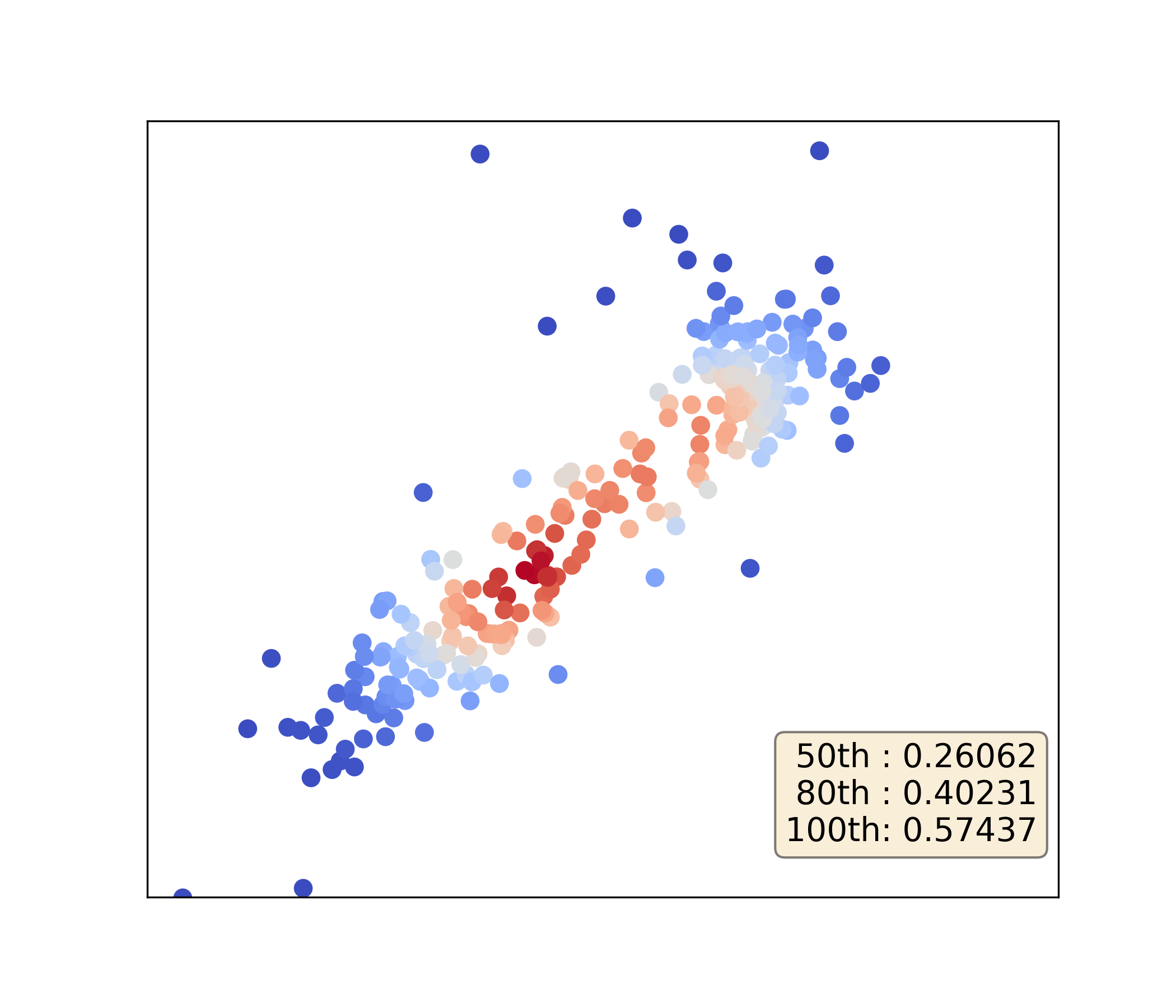}        
    \end{subfigure}    
    \hfill
    \begin{subfigure}[b]{0.32\linewidth}
        \centering
        \includegraphics[width=\linewidth]{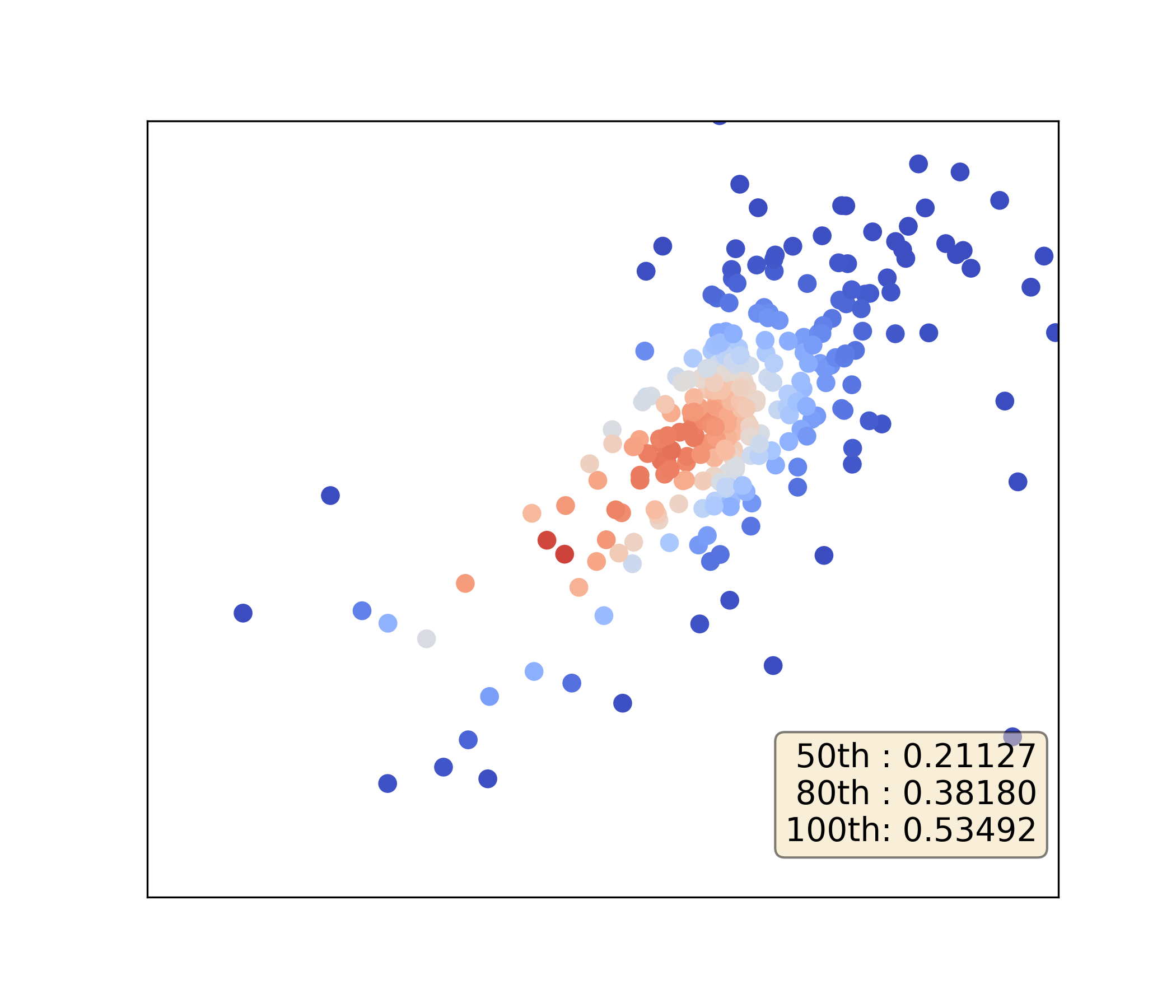}
    \end{subfigure}
    \hfill
    \begin{subfigure}[b]{0.32\linewidth}
        \centering
        \includegraphics[width=\linewidth]{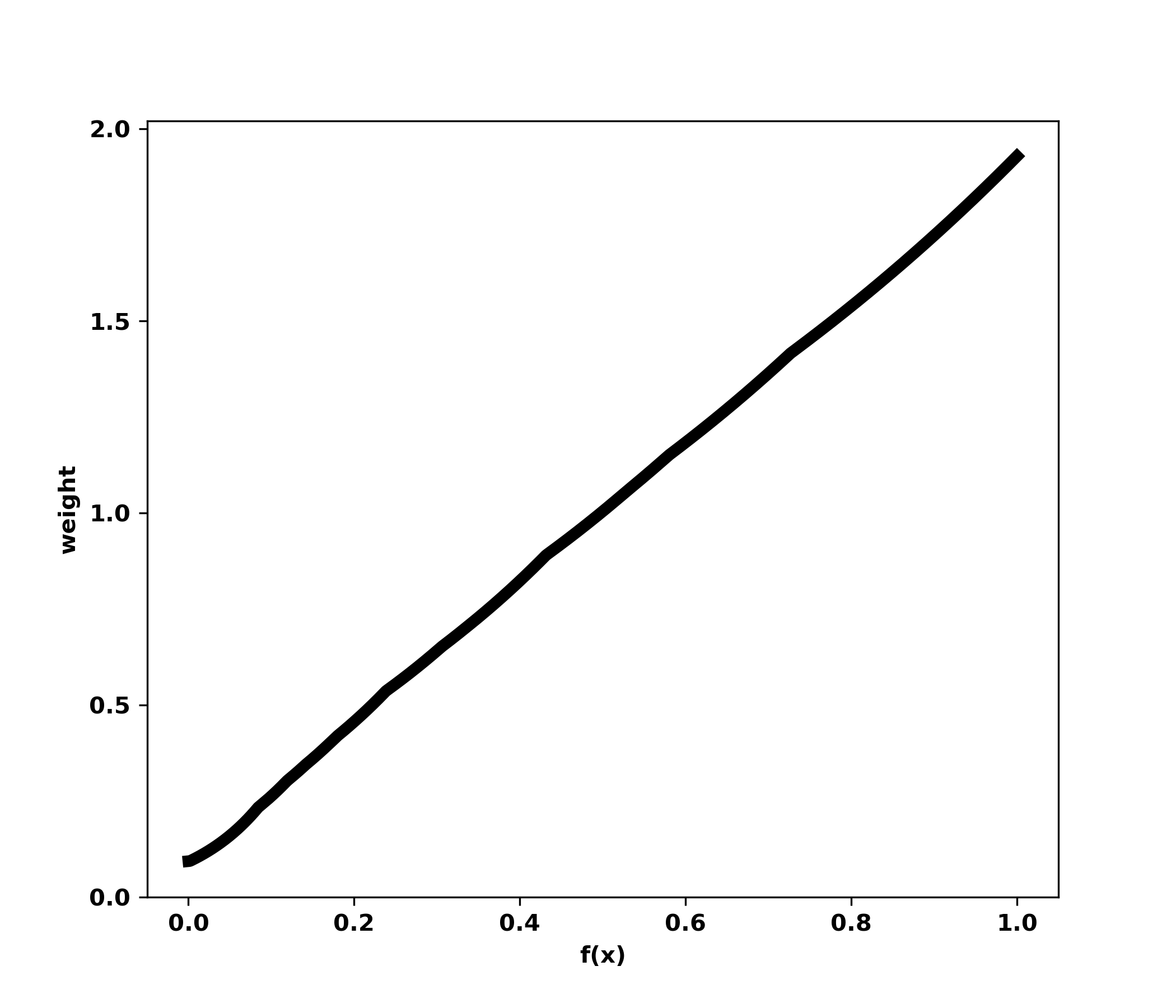}
        \caption{$\alpha=1.0$}
    \end{subfigure}
    \hfill
    \begin{subfigure}[b]{0.32\linewidth}
        \centering
        \includegraphics[width=\linewidth]{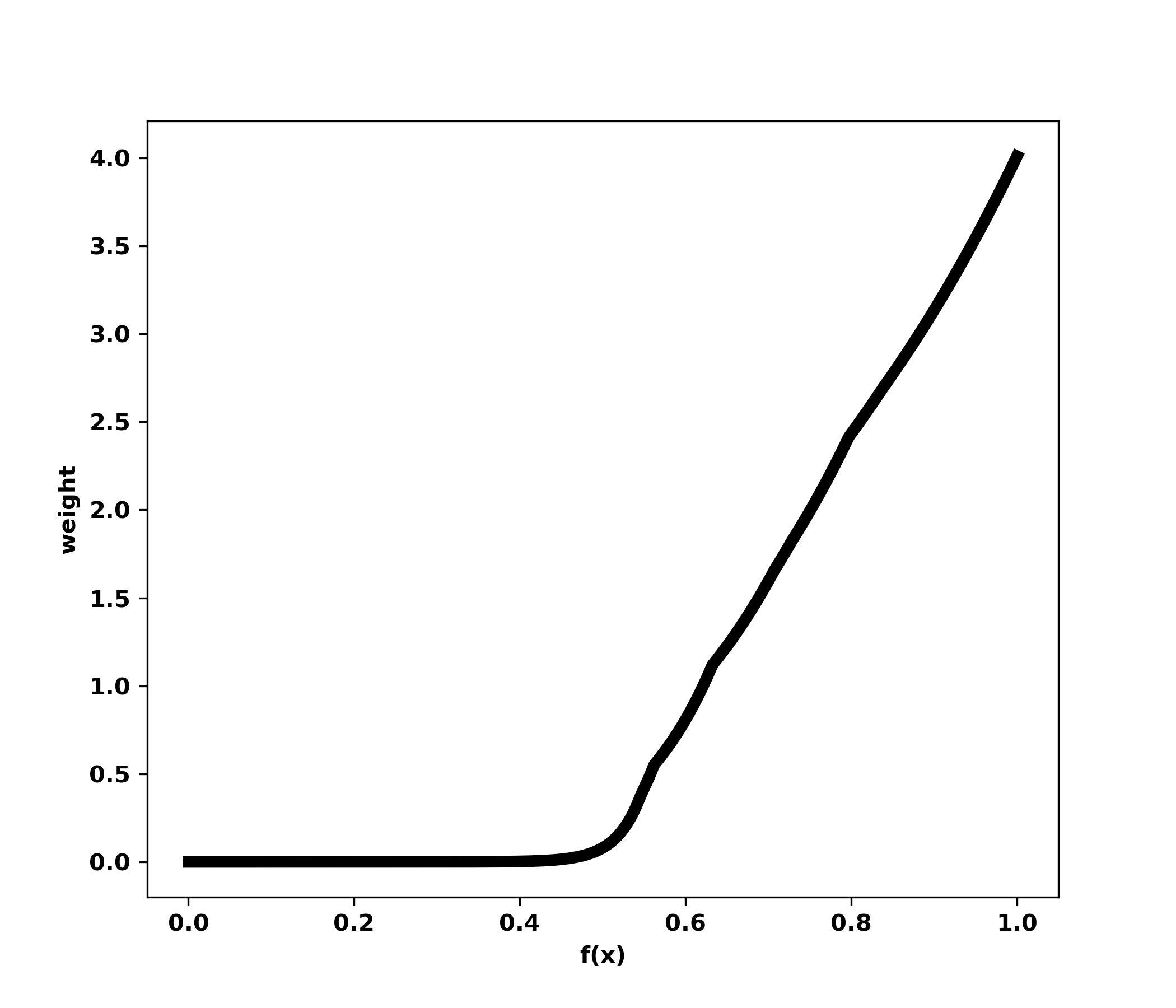}
        \caption{$\alpha=0.3$}
    \end{subfigure}
    \hfill
    \begin{subfigure}[b]{0.32\linewidth}
        \centering
        \includegraphics[width=\linewidth]{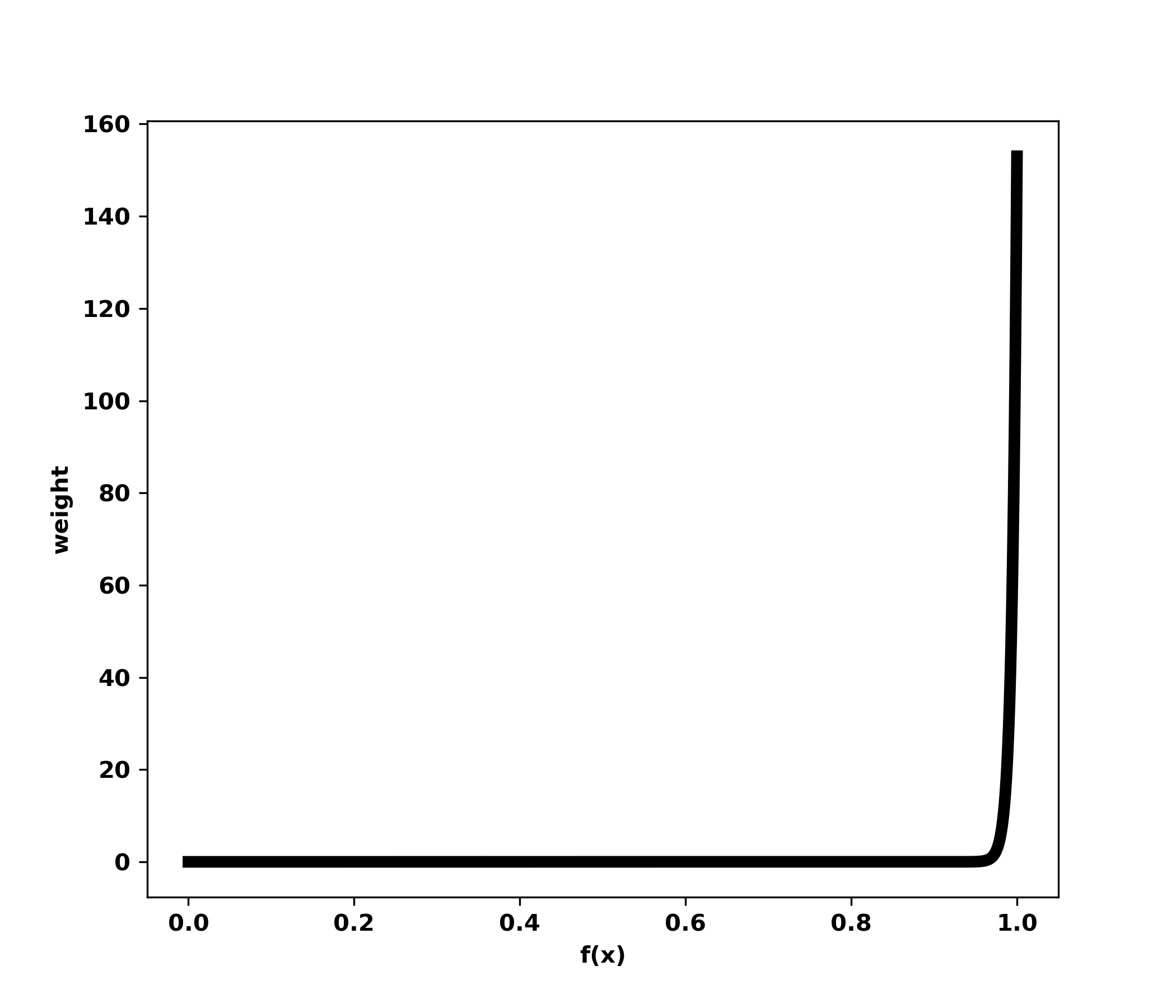}
        \caption{$\alpha=0.0$}
        \label{fig:weight}
    \end{subfigure}
    \caption{A toy example: The optimized samples and the learned weight function for different values of $\alpha$. Top: Optimized samples; Bottom: Learned weight function.}
    \label{fig:toy2}
\end{figure}

The filled contour plot of the objective function $f$ is shown in Figure~\ref{fig:contour}, with warmer colors representing higher objective values. Figure~\ref{fig:initial} shows $300$ samples drawn from the data-generating distribution $\pdata$, with the color of each sample rendered according to its ground-truth objective value. These $300$ samples and their corresponding objective values are the offline examples from which the weight function $w_\phi$ and the score-function model $\vs_\theta$ are trained. 
Figure~\ref{fig:toy2} shows the optimized samples and the learned weight function $w_\phi$ for several different values of the hyper-parameter $\alpha$ while fixing the hyper-parameter $\lambda=0.1$.

The legitimacy of the proposed approach is demonstrated by the following observations. i) Even though the weight function is \emph{not} constrained to be monotonic \emph{a priori}, as shown in Figure~\ref{fig:weight}, the learned weight functions are \emph{monotone increasing} and hence put higher weights to samples with higher objective values. ii) When $\alpha=1$, the learned weight function is relative ``flat" across its input domain. As a result, the learned generative model is very close to the data-generating distribution, and the optimized samples are very ``similar" to the initial samples. As we decrease the value of $\alpha$ from $1$ to $0.3$, the learned weight function becomes much more skewed towards the higher input values. As a result, some of the optimized samples have been nudged along the direction of the gradient of the objective function and hence have much higher objective values than the initial samples. When we further decrease the value of $\alpha$ to $0$, the learned weight function becomes extremely skewed. In this case, the hypothetical target distribution is \emph{not} learnable. As a result, instead of the gradient direction, the optimized samples have been nudged along \emph{all} directions. Therefore, the hyper-parameter $\alpha$ can effectively control the \emph{utility--learnability tradeoff} for selecting a weight function $w_\phi$. iii) Compared with the data-generating distribution, with an appropriate choice of the hyper-parameters $\alpha$ and $\lambda$, the learned generative model is substantially more \emph{capable} of generating samples with higher objective values, as demonstrated by the differences of the $50$th, $80$th, and $100$th percentiles between the samples drawn from these two distributions. More results on this toy example can be found in Appendix~\ref{sec:more toy}.

\subsection{Benchmark datasets}
Next, we assess the performance of the proposed learning algorithm using the five standard tasks (Superconductor, TF Bind 8, Ant Morphology, GFP, and UTR) from the Design-Bench benchmark \citep{trabucco2022design}. In addition, we have included the ``Fluorescence" task from \citet{fannjiang2022conformal} for a comprehensive evaluation.

\textbf{Evaluation}. We generated a total of $N = 128$ designs for each task and subsequently computed the mean and standard deviation of the $100$th percentile of the normalized ground truth over eight independent trials. The normalization process is defined by $y=(\Tilde{y} - y_{\mathrm{min}})/(y_{\mathrm{max}}-y_{\mathrm{min}})$, where $y_{\mathrm{min}}$ and $y_{\mathrm{max}}$ are chosen from the \emph{whole} dataset from which the offline examples were sampled.

\begin{table*}[!htbp]
\caption{Experimental results on the benchmark datasets [$\uparrow$].}
\label{tab:design-bench}
\begin{center}
    \begin{tabular}{c||c|c|c|c|c|c||c}
    \hline
    & Supercond. & TFBind8 & AntMorph. & GFP & UTR & Fluores. & \multirow{2}{*}{\pbox{20cm}{\small \ \ Ave. \\ Improv.}}\\
    \cline{1-7}
    \small $\mathcal{D}_{\mathrm{best}}$ & 0.399 & 0.439 & 0.565 & 0.789 & 0.593 & 0.485 \\
    \hline
    \small Grad & \textbf{0.518$\pm$\scriptsize0.024} & \textbf{0.977$\pm$\scriptsize0.025} & 0.293$\pm$\scriptsize0.023 & 0.864$\pm$\scriptsize0.001 & \textbf{0.695$\pm$\scriptsize0.013} & 0.618$\pm$\scriptsize0.204 & 0.264\\
    \small COMs & 0.439$\pm$\scriptsize0.033 & 0.945$\pm$\scriptsize0.033 & \textbf{0.944$\pm$\scriptsize0.016} & 0.864$\pm$\scriptsize0.000 & \textbf{0.699$\pm$\scriptsize0.011} & 0.588$\pm$\scriptsize0.074 & 0.402\\
    \small CbAS & \textbf{0.503$\pm$\scriptsize0.069} & 0.927$\pm$\scriptsize0.051 & 0.876$\pm$\scriptsize0.031 & \textbf{0.865$\pm$\scriptsize0.000} & \textbf{0.694$\pm$\scriptsize0.010} & 0.574$\pm$\scriptsize0.020 & 0.395\\
    \hline
    \small Ours &\textbf{0.500$\pm$\scriptsize0.051} &0.953$\pm$\scriptsize0.038 &0.844$\pm$\scriptsize0.023 &\textbf{0.865$\pm$\scriptsize0.000} &\textbf{0.698$\pm$\scriptsize0.011} &\textbf{0.721$\pm$\scriptsize0.063} & \textbf{0.446}\\
    \hline
    \end{tabular}
\end{center}
\end{table*}



\textbf{Results.} The results are listed in Table~\ref{tab:design-bench}, where $\mathcal{D}_{\mathrm{best}}$ denotes the normalized maximum objective value among the initial samples; ``Grad" refers to the vanilla gradient ascent method, in which a surrogate of the unknown objective function is learned through supervised regression; ``COMs" refers to the conservative objective models introduced in \citep{trabucco2021conservative}; ``CbAS" refers to the conditioning-by-adaptive-sampling method introduced in \citep{brookes2019conditioning}; and ``Ours" refers to the method proposed in this paper with the hyper-parameters $\alpha=0.2$ and $\lambda=0.1$. Except for the Fluorescence task, the results of existing methods are all excerpted from  \citet{trabucco2021conservative}. Results that fall within one standard deviation of the best performance are highlighted in bold. Across all tasks, our method demonstrates not only notable improvement over the best initial samples, but also \emph{consistently competitive} performances against the other three prominent offline optimization algorithms. Quantitatively, our method achieves the \emph{highest} average improvement over all six tasks, where the improvement over a specific task is defined as $(y - \mathcal{D}_{\mathrm{best}}) / \mathcal{D}_{\mathrm{best}}$. We believe that this superior consistency is rooted in our \emph{modeling} perspective and \emph{principled} design of the learning algorithm. Implementation details and additional results can be found in Appendix~\ref{sec:training details} and \ref{sec:hyperparameter}.

\section{Concluding Remarks}\label{sec:related work}
In this final section, we put the proposed \emph{PAC-generative} approach to offline optimization in the context of several related work. This will help to further elucidate the main contribution of this paper.

\textbf{Modeling target distribution \emph{vs}. modeling objective function.} As mentioned previously in Section~\ref{sec:intro}, the ``standard" approach to offline optimization is to first learn a surrogate of the unknown objective function and then apply existing optimization algorithms. The main challenge for modeling the objective function is the so-called \emph{distributional shift}. That is, when the optimization algorithm explores regions \emph{away} from the offline observations, the leaned surrogate tends to become less accurate. It is thus crucial to understand how far the optimization algorithm can explore away from the offline observations and how to maintain the accuracy of the learned surrogate throughout the exploration process. Notable effort in the literature include \citet{qi2022data} and \citet{trabucco2021conservative}, which considered regularized surrogate models in favor of invariance and conservatism; \citet{fannjiang2020autofocused} and \citet{chen2022bidirectional}, which considered surrogate models learned via importance sampling and contrastive learning; and \citet{fannjiang2022conformal}, which used conformal prediction to quantify the uncertainty of the learned surrogate. 

Despite these effort, however, it remains unclear how to align the objective of learning a surrogate of the unknown objective function with the objective of optimization. This is evidenced by the very recent work \citet{beckham2023conservative}, which discussed how one may interpret the conservative approach proposed in\citet{trabucco2021conservative}, and \citet{beckham2023exploring}, which suggested that an alternative evaluation metric is potentially better than simply choosing the best candidates using the learned surrogate. In contrast, the PAC-generative approach proposed in this paper is based on modeling a target distribution (as opposed to the objective function). As we have shown, under this generative view, it is possible to tune the objective of the learner according to a natural optimization objective. 

\textbf{Weighted learning \emph{vs}. conditional/guided generation.} Recent years have seen remarkable success in conditional/guided image generation \citep{dhariwal2021diffusion,ho2022classifier}. Conditional/guided generation can be easily adapted to the offline optimization setting. More specifically, to learn a generative model for the purpose of offline optimization, one can simultaneously learn a \emph{standard} score-based generative model and a surrogate of the objective function and then use the \emph{gradient} of the learned surrogate to guide the generation of the optimized samples \citep{maze2023diffusion}. Alternatively, one may also model the target distribution $\qtargetNP$ as the \emph{conditional} distribution $\pdata$ given $f(\rvx)\geq y_0$ for some \emph{threshold} $y_0$ and train a generative model that approximates this conditional distribution \citep{brookes2019conditioning, gupta2019feedback}. However, learning the conditional distribution $\pdata$ given $f(\rvx)\geq y_0$ may also require a surrogate of the objective function. In contrast, in our approach we model the target distribution $\qtargetNP$ using a weight function. As we have previously discussed in Section~\ref{sec:score}, in our \emph{weighted-learning} model, the score of $\qtargetNP$ is \emph{intrinsically} aligned with the gradient of the objective function. In our approach, we directly train a generative model from the offline data examples to learn the score of $\qtargetNP$, and there is \emph{no} need to learn a surrogate of the objective function separately.

\textbf{Offline optimization \emph{vs}. offline reinforcement learning (RL).} While the focus of this paper is offline optimization, recent years have also seen a substantial amount of interest in \emph{offline RL} \citep{kumar2020conservative,wang2022diffusion,janner2022planning,yuan2023reward}. Even though these two problems face some similar challenges, in our evaluation offline RL is the considerably more challenging setting. It is thus of interest to see whether the proposed PAC-generative approach can lead to any success in offline RL as well. 


\bibliographystyle{unsrtnat}
\bibliography{template}

\begin{thebibliography}{40}
\providecommand{\natexlab}[1]{#1}
\providecommand{\url}[1]{\texttt{#1}}
\expandafter\ifx\csname urlstyle\endcsname\relax
  \providecommand{\doi}[1]{doi: #1}\else
  \providecommand{\doi}{doi: \begingroup \urlstyle{rm}\Url}\fi

\bibitem[Kolli et~al.(2022)Kolli, Lu, Geng, Kumar, and Levine]{kolli2022datadriven}
Sathvik Kolli, Amy~X Lu, Xinyang Geng, Aviral Kumar, and Sergey Levine.
\newblock Data-driven optimization for protein design: Workflows, algorithms and metrics.
\newblock In \emph{ICLR2022 Machine Learning for Drug Discovery}, 2022.

\bibitem[G{\'o}mez-Bombarelli et~al.(2018)G{\'o}mez-Bombarelli, Wei, Duvenaud, Hern{\'a}ndez-Lobato, S{\'a}nchez-Lengeling, Sheberla, Aguilera-Iparraguirre, Hirzel, Adams, and Aspuru-Guzik]{gomez2018automatic}
Rafael G{\'o}mez-Bombarelli, Jennifer~N Wei, David Duvenaud, Jos{\'e}~M Hern{\'a}ndez-Lobato, Benjam{\'\i}n S{\'a}nchez-Lengeling, Dennis Sheberla, Jorge Aguilera-Iparraguirre, Timothy~D Hirzel, Ryan~P Adams, and Al{\'a}n Aspuru-Guzik.
\newblock Automatic chemical design using a data-driven continuous representation of molecules.
\newblock \emph{ACS Central Science}, 4\penalty0 (2):\penalty0 268--276, 2018.

\bibitem[Killoran et~al.(2017)Killoran, Lee, Delong, Duvenaud, and Frey]{killoran2017generating}
Nathan Killoran, Leo~J Lee, Andrew Delong, David Duvenaud, and Brendan~J Frey.
\newblock Generating and designing {DNA} with deep generative models.
\newblock \emph{arXiv preprint arXiv:1712.06148}, 2017.

\bibitem[Hoburg and Abbeel(2014)]{hoburg2014geometric}
Warren Hoburg and Pieter Abbeel.
\newblock Geometric programming for aircraft design optimization.
\newblock \emph{AIAA Journal}, 52\penalty0 (11):\penalty0 2414--2426, 2014.

\bibitem[Liao et~al.(2019)Liao, Wang, Yang, Lee, Pister, Levine, and Calandra]{liao2019data}
Thomas Liao, Grant Wang, Brian Yang, Rene Lee, Kristofer Pister, Sergey Levine, and Roberto Calandra.
\newblock Data-efficient learning of morphology and controller for a microrobot.
\newblock In \emph{2019 International Conference on Robotics and Automation}, pages 2488--2494, 2019.

\bibitem[Kumar et~al.(2022)Kumar, Yazdanbakhsh, Hashemi, Swersky, and Levine]{kumar2022datadriven}
Aviral Kumar, Amir Yazdanbakhsh, Milad Hashemi, Kevin Swersky, and Sergey Levine.
\newblock Data-driven offline optimization for architecting hardware accelerators.
\newblock In \emph{International Conference on Learning Representations}, 2022.

\bibitem[Beck(2017)]{beck2017first}
Amir Beck.
\newblock \emph{First-Order Methods in Optimization}.
\newblock SIAM, 2017.

\bibitem[Audet and Hare(2017)]{Audet2017}
Charles Audet and Warren Hare.
\newblock \emph{Introduction: Tools and Challenges in Derivative-Free and Blackbox Optimization}, pages 3--14.
\newblock Springer International Publishing, 2017.

\bibitem[Fu and Levine(2021)]{fu2021offline}
Justin Fu and Sergey Levine.
\newblock Offline model-based optimization via normalized maximum likelihood estimation.
\newblock In \emph{International Conference on Learning Representations}, 2021.

\bibitem[Brookes et~al.(2019)Brookes, Park, and Listgarten]{brookes2019conditioning}
David Brookes, Hahnbeom Park, and Jennifer Listgarten.
\newblock Conditioning by adaptive sampling for robust design.
\newblock In \emph{International Conference on Machine Learning}, pages 773--782. PMLR, 2019.

\bibitem[Gupta and Zou(2019)]{gupta2019feedback}
Anvita Gupta and James Zou.
\newblock Feedback {GAN} for {DNA} optimizes protein functions.
\newblock \emph{Nature Machine Intelligence}, 1\penalty0 (2):\penalty0 105--111, 2019.

\bibitem[Kumar and Levine(2020)]{kumar2020model}
Aviral Kumar and Sergey Levine.
\newblock Model inversion networks for model-based optimization.
\newblock \emph{Advances in Neural Information Processing Systems}, 33:\penalty0 5126--5137, 2020.

\bibitem[Trabucco et~al.(2021)Trabucco, Kumar, Geng, and Levine]{trabucco2021conservative}
Brandon Trabucco, Aviral Kumar, Xinyang Geng, and Sergey Levine.
\newblock Conservative objective models for effective offline model-based optimization.
\newblock In \emph{International Conference on Machine Learning}, pages 10358--10368. PMLR, 2021.

\bibitem[Rubinstein(1999)]{rubinstein1999cross}
Reuven Rubinstein.
\newblock The cross-entropy method for combinatorial and continuous optimization.
\newblock \emph{Methodology and Computing in Applied Probability}, 1:\penalty0 127--190, 1999.

\bibitem[Ambrosio et~al.(2021)Ambrosio, Bru{\'e}, and Semola]{Ambrosio2021}
Luigi Ambrosio, Elia Bru{\'e}, and Daniele Semola.
\newblock \emph{Lecture 3: The Kantorovich--Rubinstein Duality}, pages 23--34.
\newblock Springer International Publishing, 2021.

\bibitem[Song et~al.(2020{\natexlab{a}})Song, Sohl-Dickstein, Kingma, Kumar, Ermon, and Poole]{song2020score}
Yang Song, Jascha Sohl-Dickstein, Diederik~P Kingma, Abhishek Kumar, Stefano Ermon, and Ben Poole.
\newblock Score-based generative modeling through stochastic differential equations.
\newblock \emph{arXiv preprint arXiv:2011.13456}, 2020{\natexlab{a}}.

\bibitem[Ho et~al.(2020)Ho, Jain, and Abbeel]{ho2020denoising}
Jonathan Ho, Ajay Jain, and Pieter Abbeel.
\newblock Denoising diffusion probabilistic models.
\newblock \emph{Advances in Neural Information Processing Systems}, 33:\penalty0 6840--6851, 2020.

\bibitem[Anderson(1982)]{anderson1982reverse}
Brian~D Anderson.
\newblock Reverse-time diffusion equation models.
\newblock \emph{Stochastic Processes and their Applications}, 12\penalty0 (3):\penalty0 313--326, 1982.

\bibitem[Bogachev et~al.(2022)Bogachev, Krylov, R{\"o}ckner, and Shaposhnikov]{bogachev2022fokker}
Vladimir~I Bogachev, Nicolai~V Krylov, Michael R{\"o}ckner, and Stanislav~V Shaposhnikov.
\newblock \emph{Fokker--Planck--Kolmogorov Equations}, volume 207.
\newblock American Mathematical Society, 2022.

\bibitem[Hyv{{\"a}}rinen(2005)]{hyvarinen2005estimation}
Aapo Hyv{{\"a}}rinen.
\newblock Estimation of non-normalized statistical models by score matching.
\newblock \emph{Journal of Machine Learning Research}, 6\penalty0 (24):\penalty0 695--709, 2005.

\bibitem[Vincent(2011)]{vincent2011connection}
Pascal Vincent.
\newblock A connection between score matching and denoising autoencoders.
\newblock \emph{Neural Computation}, 23\penalty0 (7):\penalty0 1661--1674, 2011.

\bibitem[Song et~al.(2020{\natexlab{b}})Song, Garg, Shi, and Ermon]{song2020sliced}
Yang Song, Sahaj Garg, Jiaxin Shi, and Stefano Ermon.
\newblock Sliced score matching: A scalable approach to density and score estimation.
\newblock In \emph{Uncertainty in Artificial Intelligence}, pages 574--584. PMLR, 2020{\natexlab{b}}.

\bibitem[Kwon et~al.(2022)Kwon, Fan, and Lee]{kwon2022scorebased}
Dohyun Kwon, Ying Fan, and Kangwook Lee.
\newblock Score-based generative modeling secretly minimizes the wasserstein distance.
\newblock \emph{Advances in Neural Information Processing Systems}, 35:\penalty0 20205--20217, 2022.

\bibitem[Hoeffding(1994)]{hoeffding1994probability}
Wassily Hoeffding.
\newblock Probability inequalities for sums of bounded random variables.
\newblock \emph{The collected works of Wassily Hoeffding}, pages 409--426, 1994.

\bibitem[Trabucco et~al.(2022)Trabucco, Geng, Kumar, and Levine]{trabucco2022design}
Brandon Trabucco, Xinyang Geng, Aviral Kumar, and Sergey Levine.
\newblock Design-bench: Benchmarks for data-driven offline model-based optimization.
\newblock In \emph{International Conference on Machine Learning}, pages 21658--21676. PMLR, 2022.

\bibitem[Fannjiang et~al.(2022)Fannjiang, Bates, Angelopoulos, Listgarten, and Jordan]{fannjiang2022conformal}
Clara Fannjiang, Stephen Bates, Anastasios~N Angelopoulos, Jennifer Listgarten, and Michael~I Jordan.
\newblock Conformal prediction under feedback covariate shift for biomolecular design.
\newblock \emph{Proceedings of the National Academy of Sciences}, 119\penalty0 (43):\penalty0 e2204569119, 2022.

\bibitem[Qi et~al.(2022)Qi, Su, Kumar, and Levine]{qi2022data}
Han Qi, Yi~Su, Aviral Kumar, and Sergey Levine.
\newblock Data-driven offline decision-making via invariant representation learning.
\newblock \emph{Advances in Neural Information Processing Systems}, 35:\penalty0 13226--13237, 2022.

\bibitem[Fannjiang and Listgarten(2020)]{fannjiang2020autofocused}
Clara Fannjiang and Jennifer Listgarten.
\newblock Autofocused oracles for model-based design.
\newblock \emph{Advances in Neural Information Processing Systems}, 33:\penalty0 12945--12956, 2020.

\bibitem[Chen et~al.(2022)Chen, Zhang, Fu, Liu, and Coates]{chen2022bidirectional}
Can Chen, Yingxue Zhang, Jie Fu, Xue~S Liu, and Mark Coates.
\newblock Bidirectional learning for offline infinite-width model-based optimization.
\newblock \emph{Advances in Neural Information Processing Systems}, 35:\penalty0 29454--29467, 2022.

\bibitem[Beckham and Pal(2023)]{beckham2023conservative}
Christopher Beckham and Christopher Pal.
\newblock Conservative objective models are a special kind of contrastive divergence-based energy model.
\newblock \emph{arXiv preprint arXiv:2304.03866}, 2023.

\bibitem[Beckham et~al.(2023)Beckham, Piche, Vazquez, and Pal]{beckham2023exploring}
Christopher Beckham, Alexandre Piche, David Vazquez, and Christopher Pal.
\newblock Exploring validation metrics for offline model-based optimisation.
\newblock \emph{arXiv preprint arXiv:2304.03866}, 2023.

\bibitem[Dhariwal and Nichol(2021)]{dhariwal2021diffusion}
Prafulla Dhariwal and Alexander Nichol.
\newblock Diffusion models beat gans on image synthesis.
\newblock \emph{Advances in Neural Information Processing Systems}, 34:\penalty0 8780--8794, 2021.

\bibitem[Ho and Salimans(2022)]{ho2022classifier}
Jonathan Ho and Tim Salimans.
\newblock Classifier-free diffusion guidance.
\newblock \emph{arXiv preprint arXiv:2207.12598}, 2022.

\bibitem[Maz{\'e} and Ahmed(2022)]{maze2023diffusion}
Fran{\c{c}}ois Maz{\'e} and Faez Ahmed.
\newblock Topodiff: A performance and constraint-guided diffusion model for topology optimization.
\newblock \emph{arXiv preprint arXiv:2208.09591}, 2022.

\bibitem[Kumar et~al.(2020)Kumar, Zhou, Tucker, and Levine]{kumar2020conservative}
Aviral Kumar, Aurick Zhou, George Tucker, and Sergey Levine.
\newblock Conservative {Q}-learning for offline reinforcement learning.
\newblock \emph{Advances in Neural Information Processing Systems}, 33:\penalty0 1179--1191, 2020.

\bibitem[Wang et~al.(2022)Wang, Hunt, and Zhou]{wang2022diffusion}
Zhendong Wang, Jonathan~J Hunt, and Mingyuan Zhou.
\newblock Diffusion policies as an expressive policy class for offline reinforcement learning.
\newblock \emph{arXiv preprint arXiv:2208.06193}, 2022.

\bibitem[Janner et~al.(2022)Janner, Du, Tenenbaum, and Levine]{janner2022planning}
Michael Janner, Yilun Du, Joshua~B Tenenbaum, and Sergey Levine.
\newblock Planning with diffusion for flexible behavior synthesis.
\newblock \emph{arXiv preprint arXiv:2205.09991}, 2022.

\bibitem[Yuan et~al.(2023)Yuan, Huang, Ni, Chen, and Wang]{yuan2023reward}
Hui Yuan, Kaixuan Huang, Chengzhuo Ni, Minshuo Chen, and Mengdi Wang.
\newblock Reward-directed conditional diffusion: Provable distribution estimation and reward improvement.
\newblock \emph{arXiv preprint arXiv:2307.07055}, 2023.

\bibitem[Kingma and Welling(2013)]{kingma2013auto}
Diederik~P Kingma and Max Welling.
\newblock Auto-encoding variational bayes.
\newblock \emph{arXiv preprint arXiv:1312.6114}, 2013.

\bibitem[Kingma and Ba(2014)]{kingma2014adam}
Diederik~P Kingma and Jimmy Ba.
\newblock Adam: A method for stochastic optimization.
\newblock \emph{arXiv preprint arXiv:1412.6980}, 2014.

\end{thebibliography}

\clearpage
\appendix
\section{Proof of Technical Results}

\subsection{Proof of Lemma \ref{lemma}} \label{sec:lemma proof}
By assumption, we have $0 \leq \tilde{w}(\vx) \leq B$ for any $\vx \in \mathcal{X}$ and $0 \leq \ell_\theta(\vx) \leq \Delta$ for any $\vx \in \mathcal{X}$ and any $\theta \in \Theta$. It follows immediately that the \emph{weighed} loss function $\tilde{w}(\vx)\ell_{\theta}(\vx)$ satisfies $0 \leq \tilde{w}(\vx)\ell_{\theta}(\vx) \leq B\Delta$ for any $\vx \in \mathcal{X}$ and any $\theta \in \Theta$. Applying the standard Rademacher bound to the \emph{weighted} loss function class $(\tilde{w}(\vx)\ell_{\theta}(\vx):\theta\in\Theta)$, with probability $\geq 1-\delta$ we have for any $\theta\in \Theta$
\begin{align}
    \mathbb{E}_{\rvx \sim \pdata}\left[ \tilde{w}(\rvx) \ell_{\theta}(\rvx)\right] \leq \hat{\mathcal{L}}_{\vx_{[m]}}(\theta,\tilde{w})+2 \hat{\mathfrak{R}}_{\vx_{[m]}}^{\tilde{w}}(\Theta) + 3 \sqrt{\frac{ 2 B \Delta 
    \log(2/\delta) }{m}}, \label{eq:Lemma1-a}
\end{align}
where $\hat{\mathcal{L}}_{\vx_{[m]}}(\theta,\tilde{w})=\frac{1}{m}\sum_{i=1}^m\tilde{w}(\vx_i)\ell_\theta(\vx_i)$ is the empirical weighted loss over $\vx_{[m]}$, and $\hat{\mathfrak{R}}_{\vx_{[m]}}^{\tilde{w}}(\Theta)=\E_{\boldsymbol{\sigma}_{[m]}}\left[ \sup_{\theta \in \Theta} \frac{1}{m} \sum_{i=1}^m \sigma_i \tilde{w}(\vx)\ell_\theta(\vx_i)\right]$    
is the empirical \emph{weighted} Rademacher complexity with respect to the parameter family $\Theta$ over $\vx_{[m]}$. The empirical weighted Rademacher complexity $\hat{\mathfrak{R}}_{\vx_{[m]}}^{\tilde{w}}(\Theta)$ can be further bounded from above as:
\begin{eqnarray}
    \hat{\mathfrak{R}}_{\vx_{[m]}}^{\tilde{w}}(\Theta) & = & \E_{\boldsymbol{\sigma}_{[m]}}\left[ \sup_{\theta \in \Theta} \frac{1}{m} \sum_{i=1}^m \sigma_i \left(w(\vx_i)-1+1\right) \ell_\theta(\vx_i)\right] \nonumber\\
    & \leq &\E_{\sigma_{[m]}}\left[ \sup_{\theta \in \Theta} \frac{1}{m} \sum_{i=1}^m \sigma_i (\tilde{w}(\vx_i) - 1) \ell_\theta(\vx_i)\right] + \E_{\boldsymbol{\sigma}_{[m]}}\left[ \sup_{\theta \in \Theta} \frac{1}{m} \sum_{i=1}^m \sigma_i \ell_\theta(\vx_i)\right].\label{eq:Lemma1-b}
\end{eqnarray}
Further note that 
\begin{eqnarray*}
    \frac{1}{m}\sum_{i=1}^m \sigma_i (\tilde{w}(\vx_i) - 1) \ell_\theta(\vx_i) & \leq & \sqrt{\frac{\sum_{i=1}^m(\tilde{w}(\vx_i) - 1)^2}{m}}\sqrt{\frac{\sum_{i=1}^m\left(\sigma_i\ell_\theta(\vx_i)\right)^2}{m}}\\
    & = & \sqrt{\frac{\sum_{i=1}^m(\tilde{w}(\vx_i) - 1)^2}{m}}\sqrt{\frac{\sum_{i=1}^m\ell_\theta^2(\vx_i)}{m}}\\
     & \leq & \Delta\sqrt{\frac{\sum_{i=1}^m(\tilde{w}(\vx_i) - 1)^2}{m}}
\end{eqnarray*}
for any $\theta \in \Theta$ and any realization of $\boldsymbol{\sigma}_{[m]}$, where the first inequality follows from the standard Cauchy-Schwarz inequality, the second equality follows from the fact that the square of a Rademacher variable takes a constant value of $1$, and the last inequality follows from the assumption that $0 \leq \ell_\theta(\vx) \leq \Delta$ for any $\vx \in \mathcal{X}$ and any $\theta \in \Theta$. It follows immediately that
\begin{equation}
    \E_{\sigma_{[m]}}\left[ \sup_{\theta \in \Theta} \frac{1}{m} \sum_{i=1}^m \sigma_i (\tilde{w}(\vx_i) - 1) \ell_\theta(\vx_i)\right] \leq  \Delta\sqrt{\frac{\sum_{i=1}^m(\tilde{w}(\vx_i) - 1)^2}{m}}.\label{eq:Lemma1-c}
\end{equation}
Substituting (\ref{eq:Lemma1-c}) into (\ref{eq:Lemma1-b}) gives 
\begin{equation}
    \hat{\mathfrak{R}}_{\vx_{[m]}}^{\tilde{w}}(\Theta) \leq \Delta\sqrt{\hat{V}_{\vx_{[m]}}(\tilde{w})}+\hat{\mathfrak{R}}_{\vx_{[m]}}(\Theta),\label{eq:Lemma1-d}
\end{equation}
where $\hat{V}_{\vx_{[m]}}(\tilde{w})=\frac{1}{m}\sum_{i=1}^m\left(\tilde{w}(\vx_i)-1\right)^2$ is the empirical variance of $\tilde{w}$ over $\vx_{[m]}$, and $\hat{\mathfrak{R}}_{\vx_{[m]}}(\Theta)=\E_{\boldsymbol{\sigma}_{[m]}}\left[ \sup_{\theta \in \Theta} \frac{1}{m} \sum_{i=1}^m \sigma_i \ell_\theta(\vx_i)\right]$ is the empirical (un-weighted) Rademacher complexity with respect to the parameter family $\Theta$ over $\vx_{[m]}$. Substituting (\ref{eq:Lemma1-d}) into (\ref{eq:Lemma1-a}) completes the proof of (\ref{eq:Lemma1}) and hence Lemma~\ref{lemma}.

\subsection{Proof of Theorem \ref{theorem}} \label{sec:theorem proof}
For the proof, we shall write $\qtargetNP$ and $\bar{q}_{\text{target}}$ as $\qtargetNP^{\tilde{w}}$ and $\bar{q}_{\text{target}}^{\tilde{w}}$ respectively, to emphasize their dependencies on the normalized weight function $\tilde{w}$. Let us first recall from Proposition~\ref{prop:score} that the natural optimization objective $J_{\rm{opt}}(\theta)$ can be bounded from below as:
\begin{equation}
J_{\text{opt}}(\theta) \geq \mathbb{E}_{\rvx \sim \pdata} \left[
        \tilde{w}(f(\rvx))f(\rvx)\right]-c_0K\sqrt{\mathbb{E}_{\rvx \sim \pdata}\left[\tilde{w}(f(\rvx))\ell_{\text{DSM}}^{\theta}(\rvx)\right]}-c_1KW_2(\bar{q}_{\text{target}}
        ^{\tilde{w}},\mathcal{N}).
        \label{eq:theorem-a}
\end{equation}
To turn the right-hand side into a \emph{PAC} lower bound on $J_{\text{opt}}(\theta)$, let us first \emph{fix} a normalized weight function $\tilde{w} \in \tilde{\mathcal{W}}$. 

Given $\tilde{w}$, let us first apply the standard Hoeffding's inequality to obtain a \emph{concentration} lower bound on the expected utility $\mathbb{E}_{\rvx \sim \pdata} \left[\tilde{w}(f(\rvx))f(\rvx)\right]$. More specifically, by assumption we have $|f(\vx)| \leq F$ for any $\vx \in \mathcal{X}$ and $0 \leq \tilde{w}(y) \leq B$ for any $y \in [-F,F]$. It follows that the \emph{weighed} objective function $\tilde{w}(f(\vx))f(\vx)$ satisfies $|\tilde{w}(f(\vx))f(\vx)| \leq BF$ for any $\vx \in \mathcal{X}$. By Hoeffding's inequality, with probability $\geq 1-\delta'/2$ we have
\begin{equation}
\mathbb{E}_{\rvx \sim \pdata} \left[\tilde{w}(f(\rvx)) f(\rvx) \right] \geq \hat{J}_{\vx[m]}(\tilde{w}) - \sqrt{\frac{ B F \log(2/\delta') }{m}},
\label{eq:theorem-b}
\end{equation}
where $\hat{J}_{\vx[m]}(\tilde{w})=\frac{1}{m}\sum_{i=1}^m\tilde{w}(f(\vx_i))f(\vx_i)$ is the empirical utility of $\tilde{w}$. Next by Lemma~(\ref{lemma}), with probability $\geq 1-\delta'/2$ we have for \emph{any} $\theta\in\Theta$
\begin{equation*}
    \mathbb{E}_{\rvx \sim \pdata}\left[\tilde{w}(f(\rvx))\ell_{\text{DSM}}^{\theta}(\rvx)\right] \leq \hat{\mathcal{L}}_{\vx_{[m]}}(\theta,\tilde{w})+2\Delta\sqrt{\hat{V}_{\vx_{[m]}}(\tilde{w})}+2\hat{\mathfrak{R}}_{\vx_{[m]}}(\Theta) + 3 \sqrt{\frac{ 2 B \Delta 
    \log(4/\delta') }{m}}
\end{equation*}
and hence
\begin{eqnarray}
    \hspace{-20pt}\sqrt{\mathbb{E}_{\rvx \sim \pdata}\left[\tilde{w}(f(\rvx))\ell_{DSM}^{\theta}(\rvx)\right]} & \leq & \sqrt{\hat{\mathcal{L}}_{\vx_{[m]}}(\theta,\tilde{w})+2\Delta\sqrt{\hat{V}_{\vx_{[m]}}(\tilde{w})}+2\hat{\mathfrak{R}}_{\vx_{[m]}}(\Theta) + 3 \sqrt{\frac{ 2 B \Delta 
    \log(4/\delta') }{m}}}\nonumber\\
    & \leq & \sqrt{\hat{\mathcal{L}}_{\vx_{[m]}}(\theta,\tilde{w})}+\sqrt{2\Delta}\sqrt[4]{\hat{V}_{\vx_{[m]}}(\tilde{w})}+\sqrt{2\hat{\mathfrak{R}}_{\vx_{[m]}}(\Theta)} + \sqrt[4]{\frac{ 18 B \Delta 
    \log(4/\delta') }{m}},\label{eq:theorem-c}
\end{eqnarray}
where $\hat{\mathcal{L}}_{\vx_{[m]}}(\theta,\tilde{w})=\frac{1}{m}\sum_{i=1}^m\tilde{w}(\vx_i)\ell_{\text{DSM}}^\theta(\vx_i)$ 
is the empirical weighted denoising score matching loss of $\vs_\theta$ over $\vx_{[m]}$, $\hat{V}_{\vx_{[m]}}(\tilde{w})=\frac{1}{m}\sum_{i=1}^m\left(\tilde{w}(\vx_i)-1\right)^2$ is the empirical variance of $\tilde{w}$ over $\vx_{[m]}$, and $\hat{\mathfrak{R}}_{\vx_{[m]}}(\Theta)=\E_{\boldsymbol{\sigma}_{[m]}}\left[ \sup_{\theta \in \Theta} \frac{1}{m} \sum_{i=1}^m \sigma_i \ell_{\text{DSM}}^\theta(\vx_i)\right]$ is the empirical Rademacher complexity with respect to the parameter family $\Theta$ over $\vx_{[m]}$. Substituting (\ref{eq:theorem-b}) and (\ref{eq:theorem-c}) into (\ref{eq:theorem-a}), with probability $\geq 1-\delta'$ we have for \emph{any} $\theta \in \Theta$
\begin{eqnarray}
    J_{\text{opt}}(\theta) & \geq & \hat{J}_{\vx[m]}(\tilde{w}) - c_0K\sqrt{\hat{\mathcal{L}}_{\vx_{[m]}}(\theta,\tilde{w})}-c_0K\sqrt{2\Delta}\sqrt[4]{\hat{V}_{\vx_{[m]}}(\tilde{w})}-c_1KW_2(\bar{q}_{\text{target}}^{\tilde{w}
        },\mathcal{N})-\nonumber\\
    && \hspace{80pt} c_0K\sqrt{2\hat{\mathfrak{R}}_{\vx_{[m]}}(\Theta)} - c_0K\sqrt[4]{\frac{ 18 B \Delta 
    \log(4/\delta') }{m}}-\sqrt{\frac{ B F \log(2/\delta') }{m}}.\label{eq:theorem-d}
\end{eqnarray}

To remove the \emph{conditioning} on $\tilde{w}$, let $\tilde{\mathcal{W}}_\epsilon$ be an \emph{$\epsilon$-cover} of $\tilde{\mathcal{W}}$ under the $L_{\infty}$ norm. By (\ref{eq:theorem-d}), for any \emph{given} $\tilde{v}\in \tilde{\mathcal{W}}_\epsilon$, with probability $\geq 1-\delta'$ we have for \emph{any} $\theta \in \Theta$
\begin{eqnarray*}
    J_{\rm{opt}}(\theta) & \geq & \hat{J}_{\vx[m]}(\tilde{v}) - c_0K\sqrt{\hat{\mathcal{L}}_{\vx_{[m]}}(\theta,\tilde{v})}-c_0K\sqrt{2\Delta}\sqrt[4]{\hat{V}_{\vx_{[m]}}(\tilde{v})}-c_1KW_2(\bar{q}_{\text{target}}^{\tilde{v}},\mathcal{N})-\\
    && \hspace{80pt} c_0K\sqrt{2\hat{\mathfrak{R}}_{\vx_{[m]}}(\Theta)} - c_0K\sqrt[4]{\frac{ 18 B \Delta 
    \log(4/\delta') }{m}}-\sqrt{\frac{ B F \log(2/\delta') }{m}}.
\end{eqnarray*}
By the \emph{union} bound, with probability $\geq 1-|\tilde{\mathcal{W}}_\epsilon|\delta'$ we have for \emph{any} $\theta \in \Theta$ and \emph{any} $\tilde{v}\in \tilde{\mathcal{W}}_\epsilon$
\begin{eqnarray*}
    J_{\rm{opt}}(\theta) & \geq & \hat{J}_{\vx[m]}(\tilde{v}) - c_0K\sqrt{\hat{\mathcal{L}}_{\vx_{[m]}}(\theta,\tilde{v})}-c_0K\sqrt{2\Delta}\sqrt[4]{\hat{V}_{\vx_{[m]}}(\tilde{v})}-c_1KW_2(\bar{q}_{\text{target}}^{\tilde{v}},\mathcal{N})-\nonumber\\
    && \hspace{80pt} c_0K\sqrt{2\hat{\mathfrak{R}}_{\vx_{[m]}}(\Theta)} - c_0K\sqrt[4]{\frac{ 18 B \Delta 
    \log(4/\delta') }{m}}-\sqrt{\frac{ B F \log(2/\delta') }{m}}.
\end{eqnarray*}
Choosing $\delta'=\delta/|\tilde{\mathcal{W}}_\epsilon|$, with probability $\geq 1-\delta$ we have for \emph{any} $\theta \in \Theta$ and \emph{any} $\tilde{v}\in \tilde{\mathcal{W}}_\epsilon$
\begin{eqnarray}
    J_{\rm{opt}}(\theta) & \geq & \hat{J}_{\vx[m]}(\tilde{v}) - c_0K\sqrt{\hat{\mathcal{L}}_{\vx_{[m]}}(\theta,\tilde{v})}-c_0K\sqrt{2\Delta}\sqrt[4]{\hat{V}_{\vx_{[m]}}(\tilde{v})}-c_1KW_2(\bar{q}_{\text{target}}^{\tilde{v}},\mathcal{N})-\nonumber\\
    && \hspace{60pt} c_0K\sqrt{2\hat{\mathfrak{R}}_{\vx_{[m]}}(\Theta)} - c_0K\sqrt[4]{\frac{ 18 B \Delta 
    \log(4|\tilde{\mathcal{W}}_\epsilon|/\delta) }{m}}-\sqrt{\frac{ B F \log(2|\tilde{\mathcal{W}}_\epsilon|/\delta) }{m}}.\label{eq:theorem-e}
\end{eqnarray}
By the definition of $\epsilon$-cover, for any $\tilde{w}\in\tilde{\mathcal{W}}$, there exists an $\tilde{v}\in \tilde{\mathcal{W}}_\epsilon$ such that $|\tilde{w}(f(\vx))-\tilde{v}(f(\vx))| \leq \epsilon$  for any $\vx \in \mathcal{X}$. Note that this immediately implies that:
\begin{align}
    \hat{J}_{\vx[m]}(\tilde{w})-\hat{J}_{\vx[m]}(\tilde{v}) &=  \frac{1}{m}\sum_{i=1}^m \left(\tilde{w}(f(\vx_i))-\tilde{v}(f(\vx_i))\right)f(\vx_i)\nonumber\\
    & \leq \frac{1}{m}\sum_{i=1}^m \left|\tilde{w}(f(\vx_i))-\tilde{v}(f(\vx_i))\right|\left|f(\vx_i)\right| \leq F\epsilon,\label{eq:theorem-f}
\end{align}
where the last inequality follows from the assumption that $|f(\vx)| \leq F$ for any $\vx\in\mathcal{X}$;
\begin{align}
    \sqrt{\hat{\mathcal{L}}_{\vx_{[m]}}(\theta,\tilde{v})}-\sqrt{\hat{\mathcal{L}}_{\vx_{[m]}}(\theta,\tilde{w})} &= \sqrt{ \frac{1}{m}\sum_{i=1}^m \tilde{v}(f(\vx_i)) \ell_{\text{DSM}}^\theta(\vx_i)} - \sqrt{ \frac{1}{m}\sum_{i=1}^m \tilde{w}(f(\vx_i)) \ell_{\text{DSM}}^\theta(\vx_i)}\nonumber\\
    &\leq  \sqrt{\frac{1}{m}\sum_{i=1}^m\left|\tilde{v}(f(\vx_i))-\tilde{w}(f(\vx_i))\right|\ell_{\text{DSM}}^\theta(\vx_i)} \leq  \sqrt{\Delta\epsilon},
\end{align}
where the last inequality follows from the assumption that $0 \leq \ell_{\text{DSM}}^\theta(\vx) \leq \Delta$ for any $\vx \in \mathcal{X}$;
\begin{align}
    \sqrt[4]{\hat{V}_{\vx_{[m]}}(\tilde{v})}-\sqrt[4]{\hat{V}_{\vx_{[m]}}(\tilde{w})} &= \sqrt[4]{\frac{1}{m}\sum_{i=1}^m(\tilde{v}(f(\vx_i))-1)^2}-\sqrt[4]{\frac{1}{m}\sum_{i=1}^m(\tilde{w}(f(\vx_i))-1)^2}\nonumber \\
    &= \sqrt[4]{\frac{1}{m}\sum_{i=1}^m(\tilde{v}(f(\vx_i))-\tilde{w}(f(\vx_i))+\tilde{w}(f(\vx_i))-1)^2}-\sqrt[4]{\frac{1}{m}\sum_{i=1}^m(\tilde{w}(f(\vx_i))-1)^2}\nonumber \\
    & \leq \sqrt[4]{\frac{1}{m}\sum_{i=1}^m (\tilde{v}(f(\vx_i)) - \tilde{w}(f(\vx_i)) )^2}+\sqrt[4]{\frac{2}{m}\sum_{i=1}^m\left|\tilde{v}(f(\vx_i)) - \tilde{w}(f(\vx_i))\right|\left|\tilde{w}(f(\vx_i))-1\right|}\nonumber \\
    & \leq \sqrt{\epsilon} + \sqrt[4]{2(B+1)\epsilon},
\end{align}
where the last inequality follows from the assumption that $0 \leq \tilde{w}(f(\vx)) \leq B$ for any $\vx \in \mathcal{X}$; and
\begin{align}
    W_2(\bar{q}_{\text{target}}^{\tilde{v}},\mathcal{N})-W_2(\bar{q}_{\text{target}}^{\tilde{w}},\mathcal{N}) & \leq W_2(\bar{q}_{\text{target}}^{\tilde{v}}, \bar{q}_{\text{target}}^{\tilde{w}}) \nonumber\\
    & \leq c_2 W_2(q_{\text{target}}^{\tilde{v}}, q_{\text{target}}^{\tilde{w}})\nonumber\\
    & \leq c_2 d_2(\mathcal{X})d_{\text{TV}}(q_{\text{target}}^{\tilde{v}},q_{\text{target}}^{\tilde{w}})\nonumber\\
    & = \frac{1}{2}c_2d_2(\mathcal{X})\int_{\mathcal{X}} | \tilde{v}(f(\vx)) - \tilde{w}(f(\vx))| \pdata(\vx) \mathrm{d} \vx \nonumber\\
    & \leq \frac{1}{2}c_2d_2(\mathcal{X})\epsilon,\label{eq:theorem-g}
\end{align}
where $c_2$ is the \emph{Wasserstein contraction constant} of the forward process (\ref{eq:fp}),  $d_2(\mathcal{X}):=\max_{\vx,\vx'\in\mathcal{X}}\|\vx-\vx'\|_2$ is the \emph{diameter} of $\mathcal{X}$ with respect to the $\ell_2$ norm, and $d_{\text{TV}}(q_{\text{target}}^{\tilde{v}},q_{\text{target}}^{\tilde{w}})$ denotes the \emph{total variation distance} between $q_{\text{target}}^{\tilde{v}}$ and $q_{\text{target}}^{\tilde{w}}$. Here, the first inequality follows from the fact that the $2$-Wasserstein distance is a metric and hence follows the triangle inequality, the second inequality follows from the \emph{Wasserstein contraction} property of the forward process, and the third inequality follows from the \emph{total-variation} bound on the $2$-Wasserstein distance. Substituting (\ref{eq:theorem-f})--(\ref{eq:theorem-g}) into (\ref{eq:theorem-e}), with probability $\geq 1-\delta$ we have for \emph{any} $\theta \in \Theta$ and \emph{any} $\tilde{w}\in \tilde{\mathcal{W}}$
\begin{eqnarray}
    J_{\rm{opt}}(\theta) & \geq & \hat{J}_{\vx[m]}(\tilde{w}) - c_0K\sqrt{\hat{\mathcal{L}}_{\vx_{[m]}}(\theta,\tilde{w})}-c_0K\sqrt{2\Delta}\sqrt[4]{\hat{V}_{\vx_{[m]}}(\tilde{w})}-c_1KW_2(\bar{q}_{\text{target}}^{\tilde{w}},\mathcal{N})-\nonumber\\
    && \hspace{60pt} c_0K\sqrt{2\hat{\mathfrak{R}}_{\vx_{[m]}}(\Theta)} - c_0K\sqrt[4]{\frac{ 18 B \Delta 
    \log(4|\tilde{\mathcal{W}}_\epsilon|/\delta) }{m}}-\sqrt{\frac{ B F \log(2|\tilde{\mathcal{W}}_\epsilon|/\delta) }{m}}-\nonumber\\
    && \hspace{60pt} \left(F+\frac{1}{2}c_2d_2(\mathcal{X})\right)\epsilon-\left(\sqrt{\Delta}+1\right)\sqrt{\epsilon}-\sqrt[4]{2(B+1)\epsilon}. \label{eq:theorem-h}
\end{eqnarray}
By assumption, any normalized weight function from $\tilde{\mathcal{W}}$ is $L$-Lipschitz and bounded by $B$. Therefore, the covering number $|\tilde{\mathcal{W}}_\epsilon|$ is of the order $O(\exp(1/\epsilon))$. Let $\epsilon=m^{-\gamma}$ for some $\gamma\in(0,1)$, and we have from (\ref{eq:theorem-h})
\begin{eqnarray}
    J_{\rm{opt}}(\theta) & \geq & \hat{J}_{\vx[m]}(\tilde{w}) - c_0K\sqrt{\hat{\mathcal{L}}_{\vx_{[m]}}(\theta,\tilde{w})}-c_0K\sqrt{2\Delta}\sqrt[4]{\hat{V}_{\vx_{[m]}}(\tilde{w})}-c_1KW_2(\bar{q}_{\text{target}}^{\tilde{w}},\mathcal{N})-\nonumber\\
    && \hspace{60pt} c_0K\sqrt{2\hat{\mathfrak{R}}_{\vx_{[m]}}(\Theta)}-O(m^{-(1-\gamma)/4})-O(m^{-\gamma/4}). \label{eq:theorem-j}
\end{eqnarray}
Choosing $\gamma=1/2$ in (\ref{eq:theorem-j}) completes the proof of (\ref{eq:Surro2}) and hence Theorem~\ref{theorem}.

\section{Detailed Experimental Results} \label{sec:experiment details}
\subsection{Benchmark datasets} 
\label{sec:dataset details}

We conducted experiments on six standard offline optimization tasks:
\begin{itemize}
    \item \textbf{Superconductor},  which aims to design a superconductor with 86 components to maximize the critical temperature;
    \item \textbf{TF (Transcription Factor) Bind 8}, which aims to find a DNA sequence of 8 base pairs to maximize its binding affinity to a particular transcription factor;
    \item \textbf{Ant Morphology}, which aims to design the morphology of a quadrupedal ant with 60 components to enable rapid crawling;
    \item \textbf{GFP (Green Fluorescent Protein)}, which aims to find a protein sequence of 238 amino acids to maximize the fluorescence;
    \item \textbf{UTR (Untranslated Region)}, which aims to find a human 5' UTR DNA sequence of 50 base pairs to maximize the expression level of its corresponding gene;
    \item \textbf{Fluorescence}, which aims to identify a protein with high brightness. At each position, the selection of an amino acid is limited to those found in the sequences of the two parent fluorescent proteins. These parent proteins differ at precisely 13 positions in their sequences while being identical at all other positions.
\end{itemize}

For all previous tasks except for the Fluorescence, we utilized the Design-Bench package \citep{trabucco2022design} to generate the training data, pre-process the data (including the conversion of categorical features to numerical values), and evaluate new designs. For the Fluorescence task, we collected raw data from \citet{fannjiang2022conformal}. The objective value in this case is represented by the combined brightness. From a total of $2^{13} = 8192$ samples, we selected the worst 4096 examples as our training dataset. While the features in the Fluorescence dataset are binary, we simply treated them as continuous inputs to our algorithm.

\begin{table}[!t]
\caption{The benchmark datasets}
\label{tab: dataset}
\begin{center}
    \begin{tabular}{c||c|c|c|c|c|c}
    \hline
    & Supercond. & TFBind8 & AntMorph. & GFP & UTR & Fluores.\\\hline
    Type & continuous & discrete & continuous & discrete & discrete & discrete \\\hline
    Dimension & 86 & 8 & 60 & 237 & 50 & 13\\\hline
    Category & N/A & 4 & N/A & 20 & 4 & 2\\\hline
    \# Train/Total & 17014/21263 & 32898/65792 & 10004/25009 & 5000/56086 & 140k/280k & 4096/8192 \\\hline
    Min/Max & 0.0/185.0 & 0.0/1.0 & -386.9/590.2 & 1.283/4.123 & 0.0/12.0 & 0.155/1.692 \\\hline
    $\mathcal{D}_{\mathrm{best}}$ & 74.0/0.4 & 0.439/0.439 & 165.326/0.565 & 3.525/0.789 & 7.123/0.594 & 0.900/0.485 \\\hline
    \end{tabular}
\end{center}
\end{table}

The key attributes of the aforementioned benchmark datasets can be found in Table~\ref{tab: dataset}, which include:
\begin{itemize}
    \item \textbf{Type}: The type of features represented in the dataset, which can be either continuous or discrete;
    \item \textbf{Dimension}: The feature dimension of the dataset;
    \item \textbf{Category}: The number of categories for each feature (only applicable to the discrete datasets);
    \item \textbf{\# Train/Total}: The number of samples in the training and entire datasets. The entire dataset includes both the training dataset and additional data examples, which are used to help evaluate the new designs;
    \item \textbf{Min/Max}: The minimum and maximum objective values within the entire dataset;
    \item \textbf{$\mathcal{D}_{\text{best}}$}: The un-normalized and normalized maximum objective values within the training dataset.
\end{itemize}

\subsection{Implementation details} \label{sec:training details}

\textbf{Normalization.} As we adopted DDPM as our generative model, we normalized each feature to the interval $[-1, 1]$. For the objective values, we mapped the original values in the training dataset to the range of $[0, 1]$. This step ensures consistency in the learning of the (un-normalized) weight function $w_\phi$. For the GFP task, we employed a variational auto-encoder \citep{kingma2013auto} to embed the high-dimensional features into a lower-dimensional space before normalizing them into the interval $[-1,1]$.

\textbf{Networks.} In our implementation, we used neural networks to model both the (un-normalized ) weight function $w_\phi$ and the score function $\vs_t^\theta$. The weight function is a simple \emph{scalar} function. In our implementation, we simply used a 4-layer multi-layer perceptron (MLP) with ReLU activation functions. In addition, we applied an \emph{exponential} function to the output of the MLP to enforce the \emph{non-negativity} of the weight function. The architecture for the score function model consists of a time-embedding layer and five blocks of ``Dense-BatchNorm-ELU''. Before each block, we injected time-embedding information by concatenating it with the input to the block.

\textbf{Training.} The noise scheduler for the DDPM was chosen as $\beta(t)=\beta_{\text{min}}+(\beta_{\text{max}}-\beta_{\text{min}})t$ for $t\in[0,1]$, where $\beta_{\text{min}}=0.1$ and  $\beta_{\text{max}}=20$. The detailed training procedure is described in Algorithm~\ref{alg:train}. The training scheme involves first identifying a suitable \emph{initialization} of $\phi$ and $\theta$ and then followed by an \emph{alternating} maximization over $\phi$ and $\theta$. More specifically, to obtain a suitable initialization of $\phi$ and $\theta$, we first note that the model $\theta$ \emph{only} shows up in the second term of our learning objective (\ref{eq:Surro3}). Maximizing the other two terms over $\phi$ gives us an initial estimate $\phi_0$ (see Line~1 of Algorithm~~\ref{alg:train}). In our implementation, this maximization was performed via full-batch gradient descent (GD), for which we used the Adam optimizer \citep{kingma2014adam} with a constant leaning rate $10^{-3}$. Once an initial estimate $\phi_0$ has been obtained, we can obtain an initial estimate $\theta_0$ by minimizing the second term over $\theta$ while setting $\phi=\phi_0$ (see Line~2 of Algorithm~~\ref{alg:train}). To minimize the weighted denoising score matching loss, we considered a time range of $t \in [10^{-3}, 1]$ and used the Adam optimizer with a variable learning rate via stochastic gradient descent (SGD). The learning rate was gradually decreased from $10^{-3}$ to $10^{-4}$ during training. The alternating maximization of the learning objective (\ref{eq:Surro3}) over the parameters $\phi$ and $\theta$ is described in Line 3--6 of Algorithm~\ref{alg:train}. Again the Adam optimizer was used, and the learning rates were set as $\eta_1 = \eta_2 = 10^{-4}$.

\begin{algorithm}[!ht]
   \caption{\textsc{Training}}
   \label{alg:train}
   \begin{algorithmic}
   \State \textbf{Input}: Offline examples $\left((\vx_i, f(\vx_i)): i \in [m] \right)$; hyper-parameters $\alpha,\lambda$; learning-rate parameters $\eta_1,\eta_2$.
   \State \textbf{General step}:
   \begin{itemize}
        \item[1:] $\phi_0 \leftarrow \argmax_{\phi \in \Phi } \left\{\frac{1}{m}\sum_{i=1}^m \frac{w_{\phi}(f(\vx_i))f(\vx_i)}{\hat{Z}_{\phi}}-\alpha\sqrt[4]{\frac{1}{m}\sum_{i=1}^m\left(\frac{w_\phi(f(\vx_i))}{\hat{Z}_{\phi}}-1\right)^2} \right\}$\Comment{via GD}
        \item[2:] $\theta_0 \leftarrow \argmin_{\theta \in \Theta} \left\{ \frac{1}{m}\sum_{i=1}^m\frac{w_{\phi_0}(f(\vx_i)) \cdot \ell_{\text{DSM}}^{\theta}(\vx_i)} {\hat{Z}_{\phi_0}} \right\}$\Comment{via SGD}
        \item[3:] \textbf{for} $k = 0, 1, \ldots, K-1,$ \textbf{do}
        \item[4:] $\qquad \phi_{k+1} \leftarrow \phi_{k} + \eta_1 \cdot \nabla_{\phi} J_{\alpha,\lambda}(\theta_{k}, \phi_{k})$\Comment{via GD}
        \item[5:] $\qquad \theta_{k+1} \leftarrow  \theta_{k} - \eta_2 \cdot \nabla_\theta \left\{ \frac{1}{m}\sum_{i=1}^m\frac{w_{\phi_{k+1}}(f(\vx_i)) \cdot \ell_{\text{DSM}}^{\theta_{k}}(\vx_i)} {\hat{Z}_{\phi_{k+1}}} \right\}$\Comment{via SGD}
        \item[6:] \textbf{end for}
   \end{itemize}
   \State \textbf{Output}: Model parameters $(\phi^*,\theta^*) = (\phi_K,\theta_{K})$.
   \end{algorithmic}
\end{algorithm} 

\textbf{Sampling/Optimization.}
The sampling/optimization procedure is described in Algorithm~\ref{alg:opt}. This procedure is identical to the probability-flow sampler in \citet{song2020score}.

\begin{algorithm}[!ht]
   \caption{\textsc{Sampling/Optimization}}
   \label{alg:opt}
   \begin{algorithmic}
   \State \textbf{Input}: Score function model $\vs_t^{\theta^{*}}(\vx)$, number of samples $N$, number of steps $T$, noise scheduler parameters $(\beta_{\mathrm{min}}, \beta_{\mathrm{max}})$, and $\Tilde{\beta}(t) = \frac{1}{T} \left[ \beta_{\mathrm{min}} +  \frac{t}{T} (\beta_{\mathrm{max}} - \beta_{\mathrm{min}}) \right]$.
   \State \textbf{General step}: 
   \begin{itemize}
       \item[1:] Draw $N$ samples $\vx_T^{(1)}, \vx_T^{(2)}, \ldots, \vx_T^{(N)} \stackrel{\text{i.i.d.}}{\sim} \mathcal{N}(0, \mI)$
       \item[2:] \textbf{for} $n = 1, 2, \ldots, N$, \textbf{do} 
       \item[3:] \qquad \textbf{for} $t = T, T-1, \ldots, 1,$ \textbf{do} 
       \item[4:] \qquad \qquad $\vx_{t-1}^{(n)} \leftarrow \left(2-\sqrt{1-\Tilde{\beta}(t)} \right) \cdot \vx_{t}^{(n)} + \frac{1}{2} \Tilde{\beta}(t) \cdot \vs_{t/T}^{\theta^*}(\vx_t^{(n)})$
       \item[5:] \qquad \textbf{end for}
       \item[6:] \textbf{end for}
   \end{itemize}
   \State \textbf{Output}: Optimized samples $\vx_0^{(1)}, \vx_0^{(2)}, \ldots, \vx_0^{(N)}$.
   \end{algorithmic}
\end{algorithm} 

\subsection{Additional experimental results}
\subsubsection{Toy example} \label{sec:more toy}
\begin{figure}[!htbp]
    \centering 
    \includegraphics[width=\textwidth]{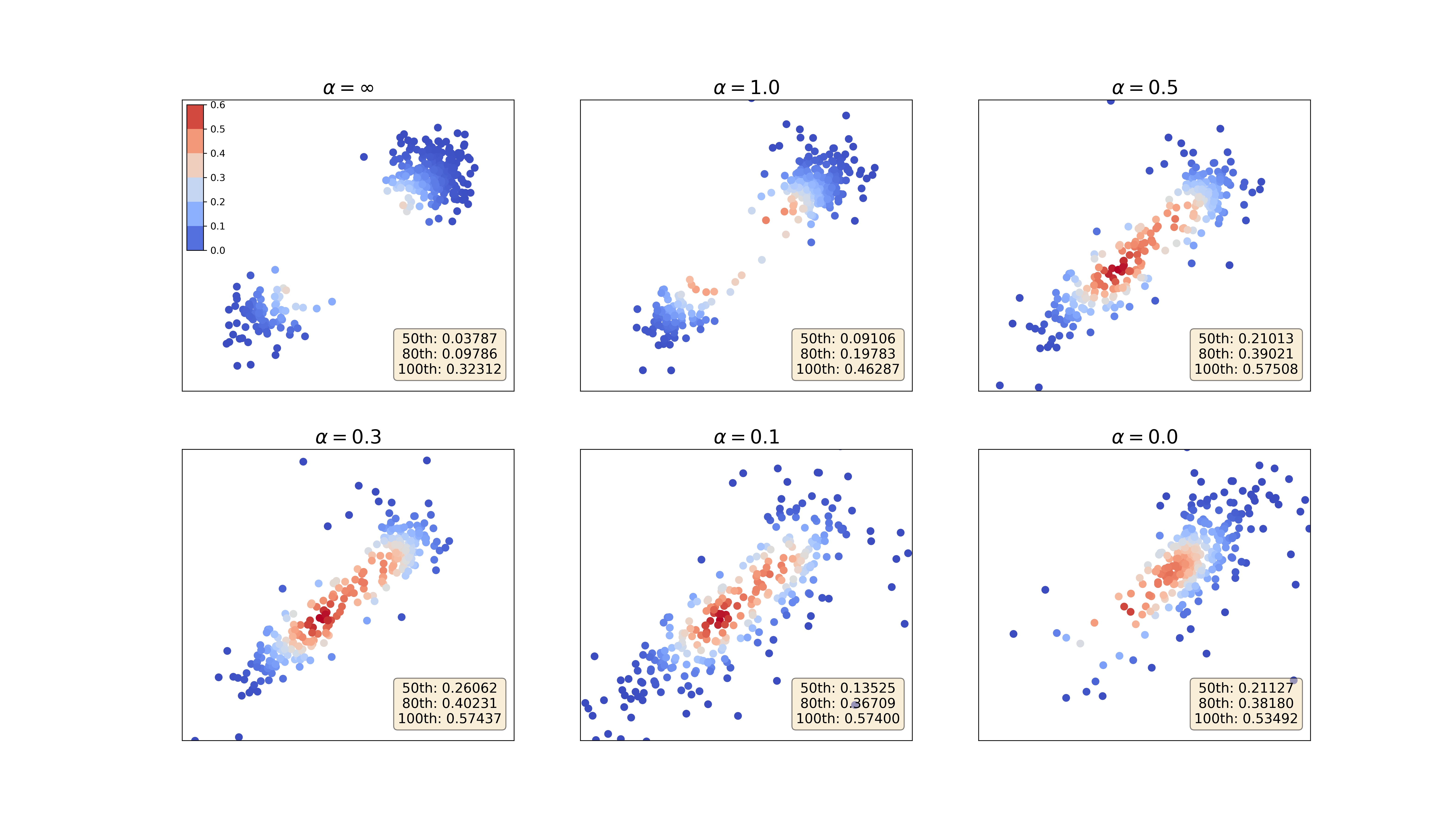}
    \caption{Optimized samples (with trainable weight function) for different choices of the hyper-parameter $\alpha$.}
    \label{fig:toy3}
\end{figure}

\begin{figure}[!htbp]
    \centering
    \includegraphics[width=\textwidth]{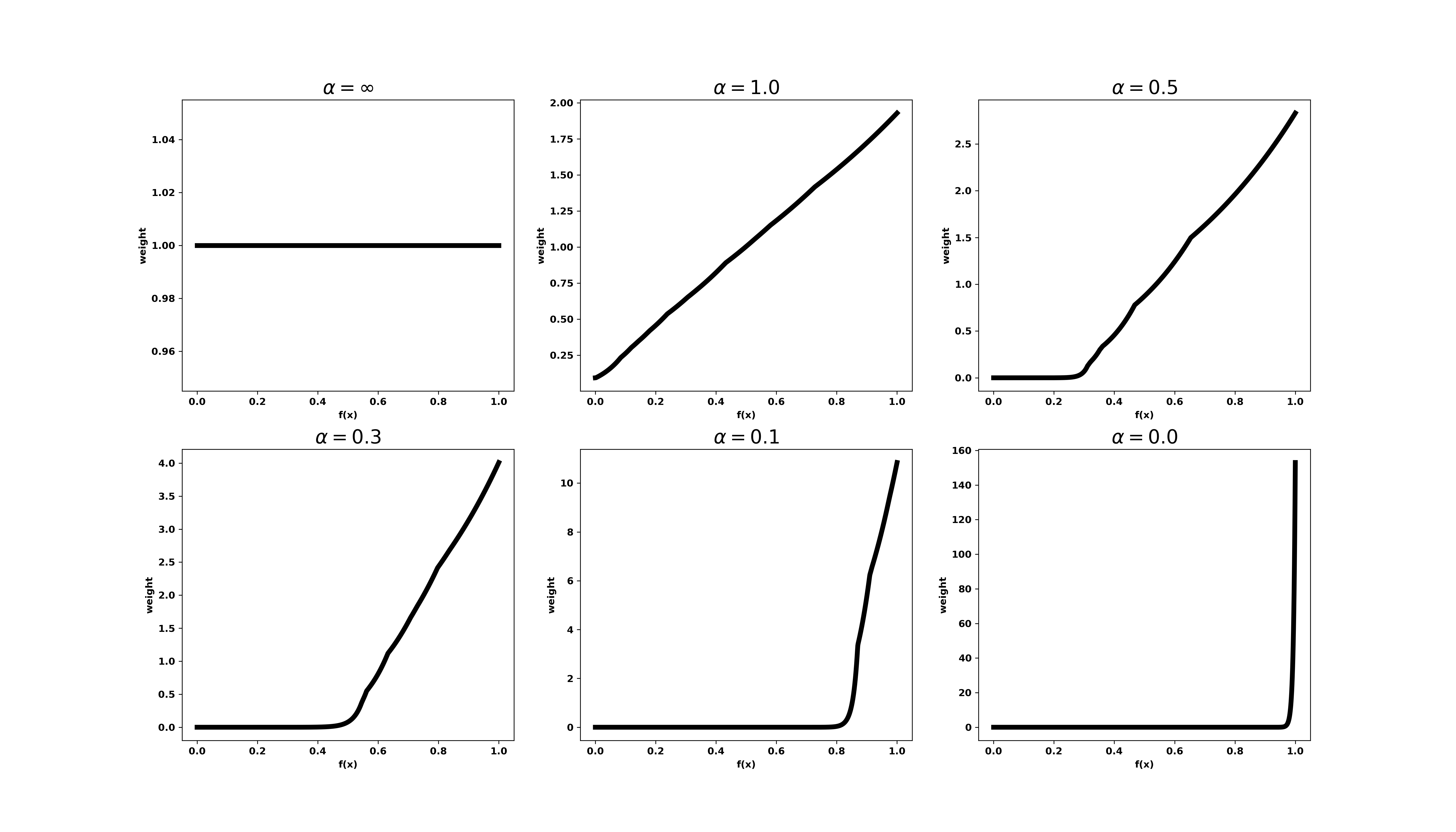}
    \caption{Learned Weight function $w_{\phi^*}$ for different choices of the hyper-parameter $\alpha$.}
    \label{fig:toy4}
\end{figure}

\textbf{Additional choices of the hyper-parameter $\alpha$.} Previously in Section~\ref{sec:toy example}, we described a toy example in $\mathbb{R}^2$ and used it to validate our proposed approach. In particular, in Figure~\ref{fig:toy2} we reported the optimized samples and the learned weight function $w_{\phi^*}$ for several choices of the hyper-parameter $\alpha$. Here in Figures~\ref{fig:toy3} and \ref{fig:toy4} we report the optimized samples and the learned weight function $w_{\phi^*}$ for some additional choice of the hyper-parameter $\alpha$. Note that when $\alpha=\infty$, the learned weight function $w_{\phi^*}$ is completely flat across its domain, and thus the hypothetical target distribution $\qtargetNP$ is identical to the data-generating distribution $\pdata$. It should become very clear from these reported results that the hyper-parameter $\alpha$ can effectively control the utility-learnability tradeoff for selecting a weight function.

\begin{figure}[!htbp]
    \centering 
    \includegraphics[width=\textwidth]{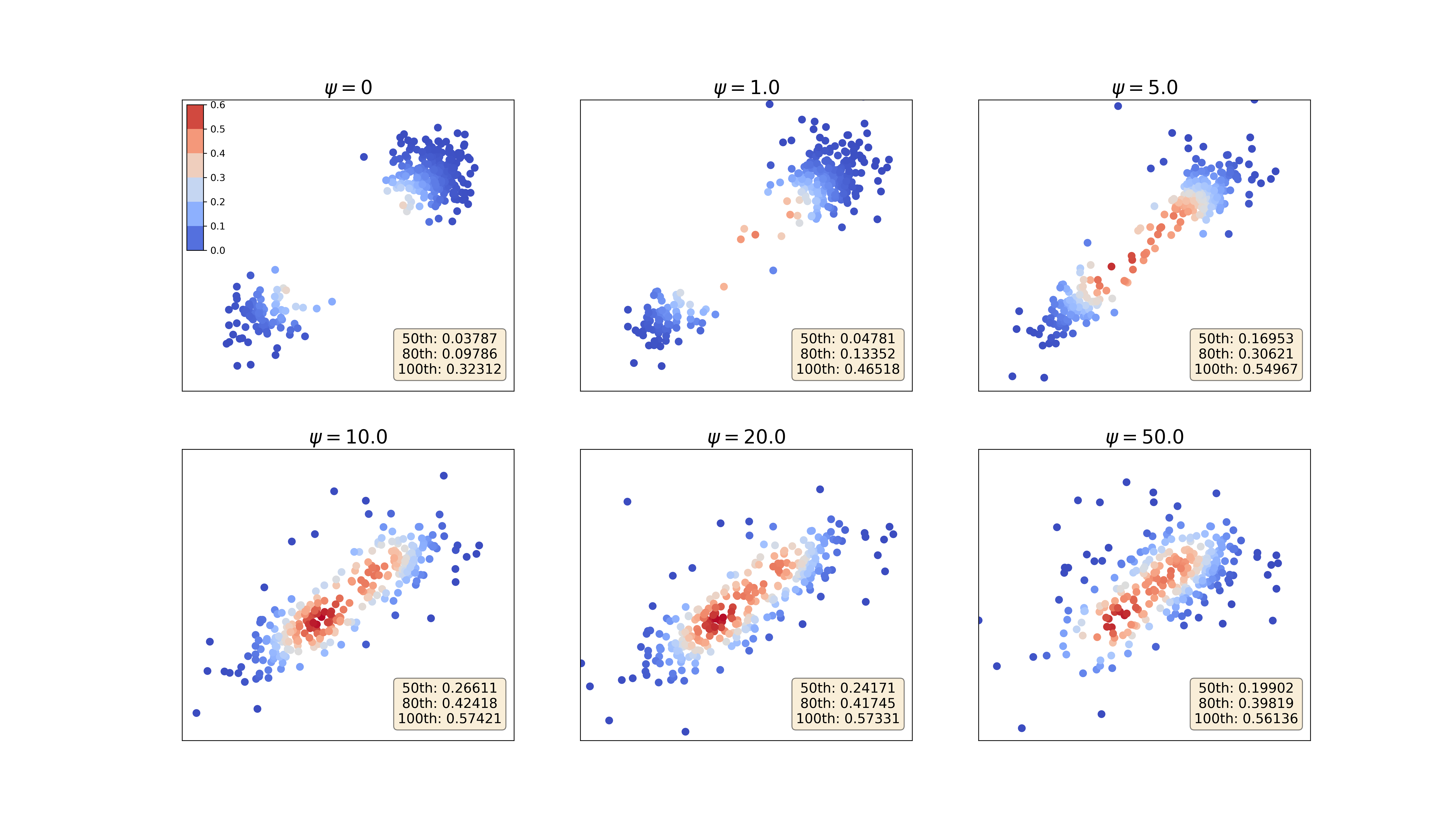}
    \caption{Optimized samples with predefined weight function for different choices of the hyper-parameter $\psi$.}
    \label{fig:toy5}
\end{figure}

\begin{figure}[!htbp]
    \centering
    \includegraphics[width=\textwidth]{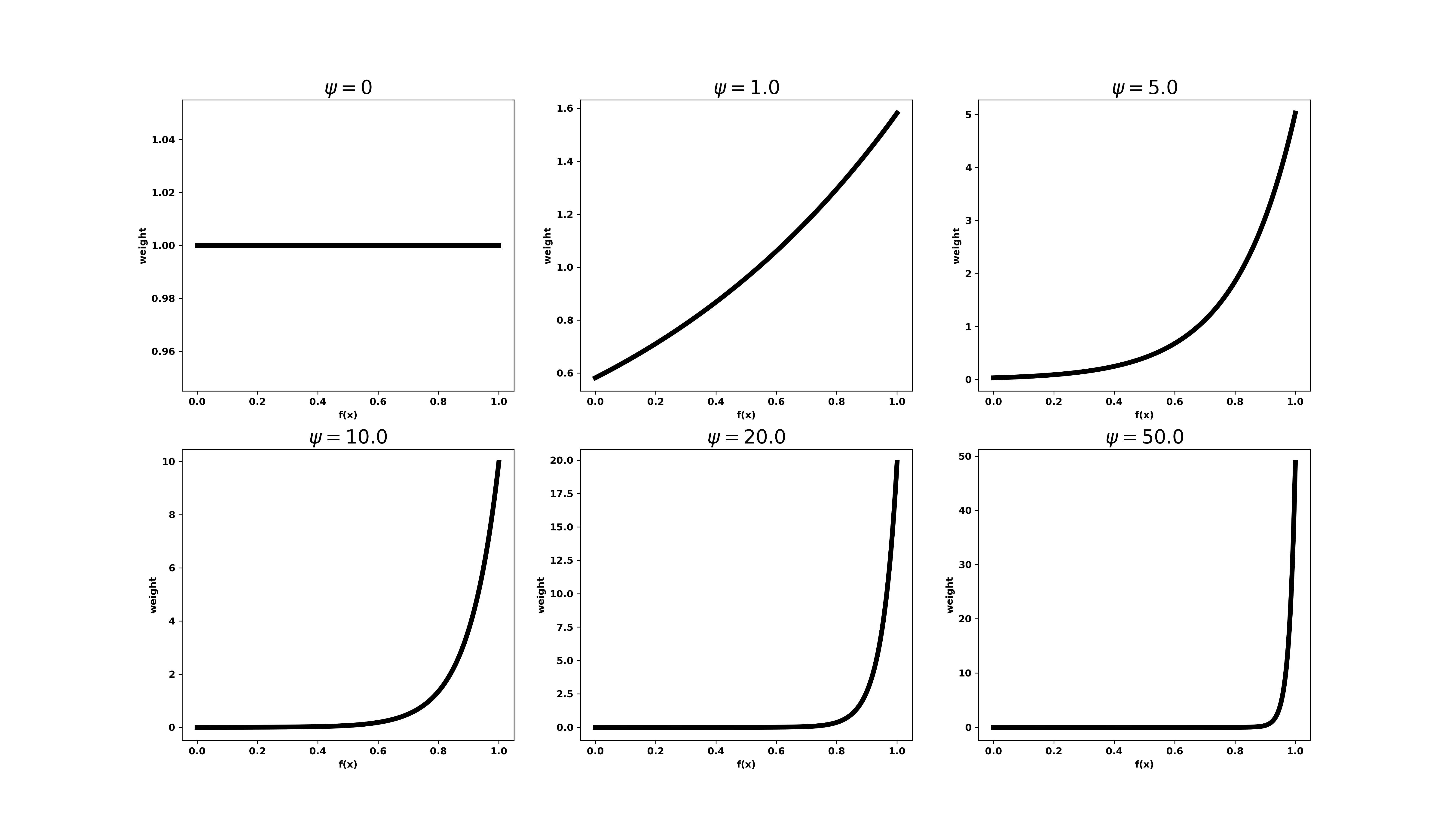}
    \caption{Predefined weight function $w_\psi$ for different choices of the hyper-parameter $\psi$.}
    \label{fig:toy6}
\end{figure}

\textbf{Predefined weight function.} Instead of considering a \emph{trainable} weight function $w_\phi$, we may also consider using a \emph{predefined} weight function to train the generative model $\ptargetNP$. Motivated by the learned weight functions $w_{\phi^*}$ from Figure~\ref{fig:toy4}, here we consider the simple \emph{exponential} function $w_\psi(y)=\exp(\psi y)$, where $\psi$ is a hyper-parameter. Note that when $\psi=0$, the weight function $w_\psi$ is completely flat across its domain, and as we increase the value of $\psi$, $w_\psi$ becomes increasingly skewed towards the higher values in its domain. The optimized samples and the corresponding predefined weight functions are reported in Figures~\ref{fig:toy5} and \ref{fig:toy6}. Note here that we have purposely chosen the values of the hyper-parameter $\psi$ such that the predefined weight functions $w_\psi$ in Figure~\ref{fig:toy6} \emph{mimic} the learned weight function $w_{\phi^*}$ in Figure~\ref{fig:toy4}. As a result, the optimized samples from Figures~\ref{fig:toy5} have similar statistical profiles as those from Figures~\ref{fig:toy3}. Next, we shall use the benchmark datasets to illustrate that a trainable weight function can significantly outperform a predefined weight function in terms of generating samples with a consistent and superior statistical profile.

\subsubsection{Benchmark datasets} \label{sec:hyperparameter} 
\begin{figure}[!t]
    \centering
    \includegraphics[width=\textwidth]{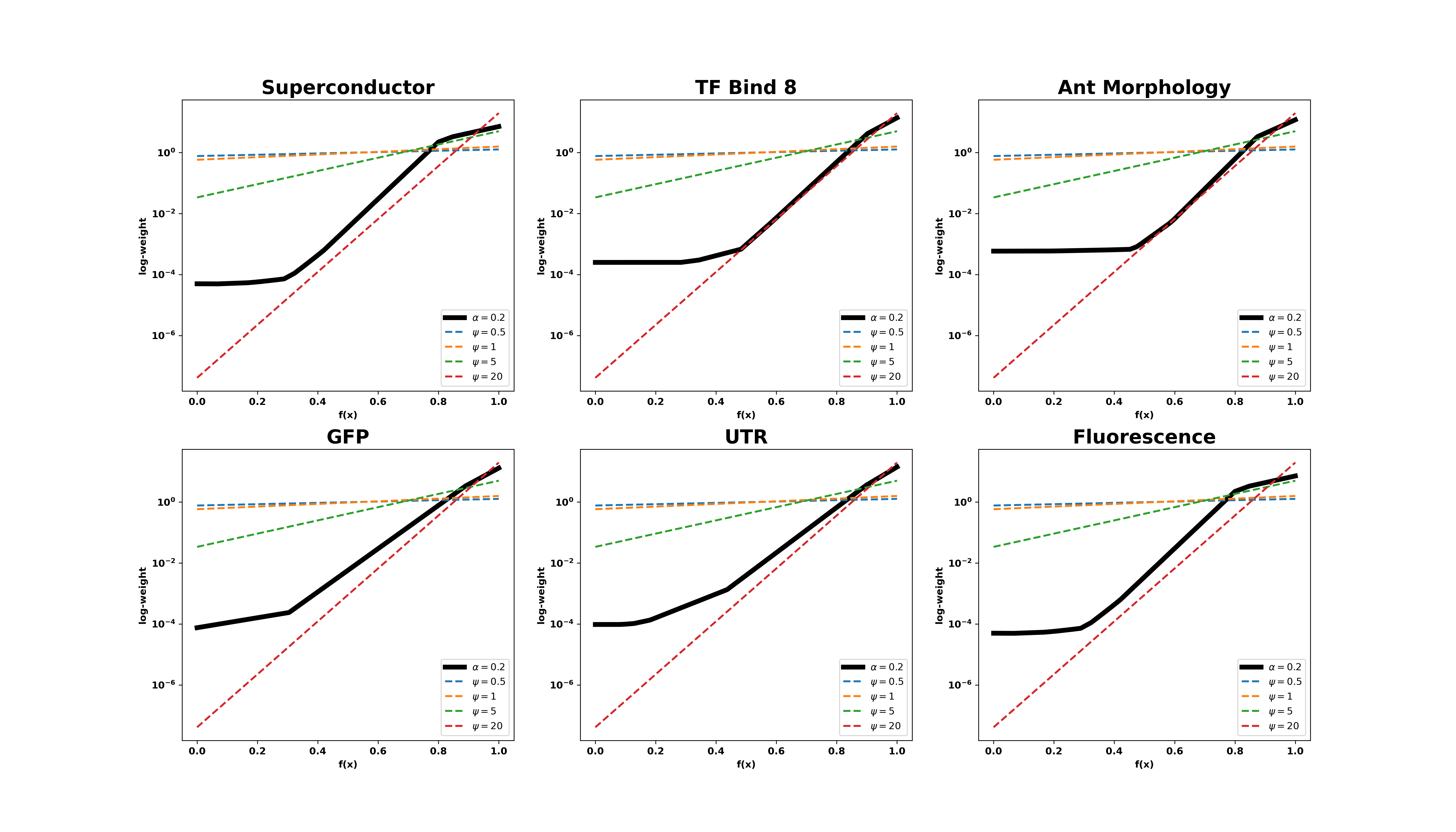}
    \caption{Learned weight functions $w_{\phi^*}$ for the benchmark datasets.}
    \label{fig:ant weight}
\end{figure}

Here we report additional results on the benchmark datasets using both the trainable weight function $w_\phi$ and the predefined exponential weight function $w_\psi$. In our experiments, we fixed the value of the hyper-parameter $\lambda=0.1$ and considered several different values for the hyper-parameter $\alpha$ (trainable weight function) and $\psi$ (predefined weight function). The mean and standard deviation of the \emph{best} generated samples for each benchmark dataset are reported in Table~\ref{tab:compare weight function}. The average improvements for different choices of the hyper-parameter $\alpha$ (trainable weight function) and $\psi$ (predefined weight function) are reported in Table~\ref{tab:ave improvement}. (All values reported in Tables~\ref{tab:compare weight function} and \ref{tab:ave improvement} are based on \emph{un-normalized} objective values.) It is clear that the use of a trainable weight function with $\alpha=0.2$ significantly outperform any predefined weight function considered here in terms of the average improvement. The learned weight functions $w_{\phi^*}$ that correspond to $\alpha=0.2$ for each of the benchmark datasets are reported in Figure~\ref{fig:ant weight}. Clearly, for the benchmark datasets, the learned weight functions are substantially different from the exponential predefined weight functions. To unleash the true power of the generative approach to offline optimization, it is thus critical to make the weight function trainable as well.

\begin{table}[!ht]
\caption{Mean and standard deviation of the best generated samples for different choices of the hyper-parameter $\alpha$ (trainable weight function) and $\psi$ (predefined weight function)
}
\label{tab:compare weight function}
\begin{center}
    \makebox[\textwidth]{
    \begin{tabular}{c||c|c|c|c|c|c}
    \hline
    & Supercond. & TFBind8 & AntMorph. & GFP & UTR & Fluores.\\
    \hline
    $\mathcal{D}_{\mathrm{best}}$ & 74 & 0.439 & 165.326 & 3.525 & 7.061 & 0.900 \\
    \hline
    \rowcolor{lightgray} $\alpha=0.15$ &75.251$\pm$11.186 &0.939$\pm$0.053 &397.651$\pm$25.994 &3.739$\pm$0.001 &8.328$\pm$0.088 &1.312$\pm$0.092 \\
    $\alpha=0.2$ &92.570$\pm$9.398 &0.953$\pm$0.038 &437.968$\pm$22.148 &3.739$\pm$0.001 &8.380$\pm$0.128 &1.263$\pm$0.097 \\
    \rowcolor{lightgray} $\alpha=0.25$ &98.848$\pm$8.940 &0.929$\pm$0.032 &415.724$\pm$31.983 &3.738$\pm$0.001 &8.390$\pm$0.111 &1.335$\pm$0.065 \\
    $\alpha=0.3$ &90.352$\pm$5.624 &0.922$\pm$0.056 &402.858$\pm$61.336 &3.739$\pm$0.001 &8.369$\pm$0.127 &1.323$\pm$0.103 \\
    \hline\hline
    \rowcolor{lightgray} $\psi=0.5$ &84.509$\pm$4.248 &0.930$\pm$0.051 &362.179$\pm$51.306 &3.739$\pm$0.001 &7.969$\pm$0.162 &1.465$\pm$0.069 \\
    $\psi=1.0$ &91.931$\pm$6.699 &0.893$\pm$0.031 &325.462$\pm$73.974 &3.739$\pm$0.001 &8.030$\pm$0.114 &1.443$\pm$0.137 \\
    \rowcolor{lightgray} $\psi=5.0$ &91.206$\pm$7.440 &0.887$\pm$0.064 &381.254$\pm$48.028 &3.739$\pm$0.001 &8.390$\pm$0.179 &1.436$\pm$0.068 \\
    $\psi=20.0$ &75.589$\pm$7.690 &0.942$\pm$0.045 &392.044$\pm$63.137 &3.739$\pm$0.001 &8.336$\pm$0.078 &1.325$\pm$0.078 \\
    \hline
    \end{tabular}}
\end{center}
\end{table}

\begin{table}[!ht]
\caption{Average improvement for different choices of the hyper-parameter $\alpha$ (trainable weight function) and $\psi$ (predefined weight function)
}
\label{tab:ave improvement}
\begin{subtable}[!ht]{0.49\textwidth}
\caption{Trainable weight function}
    \begin{center}
    \begin{tabular}{c|c}
    \hline
    $\alpha$ &Ave. Improvement\\ 
    \hline
    \rowcolor{lightgray} $0.15$ &0.543  \\
    \hline
    $0.2$ &0.620 \\
    \hline
    \rowcolor{lightgray} $0.25$ &0.616  \\
    \hline
    $0.3$ &0.579  \\
    \hline
    \end{tabular}
    \end{center}
\end{subtable}
\hfill
\begin{subtable}[!ht]{0.49\textwidth}
\caption{Predefined weight function}
    \begin{center}
    \begin{tabular}{c|c}
    \hline
    $\psi$ &Ave. Improvement \\ 
    \hline
    \rowcolor{lightgray} $0.5$ &0.545 \\
    \hline
    $1.0$ &0.508 \\
    \hline
    \rowcolor{lightgray} $5.0$ &0.567 \\
    \hline
    $20.0$ &0.542 \\
    \hline
    \end{tabular}
    \end{center}
\end{subtable}
\end{table}

\end{document}